\documentclass[journal]{IEEEtran}%[color=green]
\usepackage[final]{changes}%  [color=green]       修订模式 [color=green] 最终输出版本，[final]不显示修订  \usepackage[final]{changes}
\definechangesauthor[name={Author1}, color=red]{Reviewer1}
\definechangesauthor[name={Author2}, color=blue]{Reviewer2}
\definechangesauthor[name={Author3}, color=yellow]{Reviewer3}
\definechangesauthor[name={Author4}, color=orange]{Reviewer4}
\usepackage[para]{threeparttable}
\usepackage{amssymb,amsmath}

\usepackage{diagbox}
\usepackage[pagebackref=false,colorlinks]{hyperref}
\usepackage{cite}
\usepackage{hyperref}  
\usepackage{amsmath} 
\usepackage{booktabs}
\usepackage{multirow}
\usepackage{caption}%  
\usepackage{microtype}% 
\usepackage{verbatim}% 
\usepackage{bm} %
\usepackage{indentfirst}%
\usepackage{graphicx}
\usepackage{float}
\usepackage{rotate}
\usepackage[section]{placeins}  %  
\usepackage{setspace}  %  
\usepackage{paralist}
\usepackage{subfigure}

\usepackage{makecell} % 
\usepackage{setspace}  %  

\usepackage[switch]{lineno}

\title{ HQDec: Self-Supervised Monocular Depth Estimation Based on a High-Quality Decoder} 
\author{Fei~Wang,~\IEEEmembership{Student Member,~IEEE}, Jun~Cheng,~\IEEEmembership{Member,~IEEE}
\thanks{\newline\indent The manuscript, which is currently still under review, was submitted to IEEE Transactions on Circuits and Systems for Video Technology on December 20, 2022. Corresponding author: Jun~Cheng. \newline\indent Fei~Wang, Jun~Cheng are with the Guangdong-Hong Kong-Macao Joint Laboratory of Human-Machine Intelligence-Synergy Systems, Shenzhen Institute of Advanced Technology, Chinese Academy of Sciences, Shenzhen, 518055, China. Fei~Wang is also with University of Chinese Academy of Sciences, Beijing,100049, China. They are also with The Chinese University of Hong Kong (email:\{fei.wang2, jun.cheng\}@siat.ac.cn).}}

\usepackage{bbding}

\let\oldequation\equation
\let\oldendequation\endequation

\renewenvironment{equation}
{\linenomathNonumbers\oldequation}
{\oldendequation\endlinenomath}

\begin{document}
\maketitle
\begin{abstract}

Decoders play significant roles in recovering scene depths. However, the decoders used in previous works ignore the propagation of multilevel lossless fine-grained information, cannot adaptively capture local and global information in parallel, and cannot perform sufficient global statistical analyses on the final output disparities. In addition, the process of mapping from a low-resolution feature space to a high-resolution feature space is a one-to-many problem that may have multiple solutions. Therefore, the quality of the recovered depth map is low. To this end, we propose a high-quality decoder (HQDec), with which multilevel near-lossless fine-grained information, obtained by the proposed adaptive axial-normalized position-embedded channel attention sampling module (AdaAxialNPCAS), can be adaptively incorporated into a low-resolution feature map with high-level semantics utilizing the proposed adaptive information exchange scheme. In the HQDec, we leverage the proposed adaptive refinement module (AdaRM) to model the local and global dependencies between pixels in parallel and utilize the proposed disparity attention module to model the distribution characteristics of disparity values from a global perspective. To recover fine-grained high-resolution features with maximal accuracy, we adaptively fuse the high-frequency information obtained by constraining the upsampled solution space utilizing the local and global dependencies between pixels into the high-resolution feature map generated from the nonlearning method. Extensive experiments demonstrate that each proposed component improves the quality of the depth estimation results over the baseline results, and the developed approach achieves state-of-the-art results on the KITTI and DDAD datasets. The code and models will be publicly available at \href{https://github.com/fwucas/HQDec}{HQDec}.

\end{abstract}
\IEEEpeerreviewmaketitle

\vspace{-0.5cm}
\section{Introduction}\label{sec:intro}

\IEEEPARstart {D}{epth} plays a crucial role in mobile robot vision and navigation \cite{yasuda2020autonomous}, smart medical information technology \cite{liu2019dense,shao2021self,shao2022self}, and industrial robots \cite{zeng2017multi}. Although the existing traditional methods \cite{eigen2014depth,liu2015learning,fu2018deep,jiao2018look,wang2020cliffnet,miangoleh2021boosting,huynh2020guiding,parida2021beyond,bhat2021adabins,ranftl2021vision,yang2021transformer,lee2021patch,lee2022edgeconv,li2021revisiting,park2019high,meng2021cornet,song2021monocular} can achieve competitive results, these methods require expensive depth sensors and considerable labor to obtain sufficient data labeled with pixel-level depth information or even require stereo video sequences \cite{li2021revisiting,park2019high} for network training. Labeled depth data are expensive to acquire and can only be applied to limited scenarios. To alleviate this constraint, researchers have recently attempted to directly infer depths from large amounts of easily accessible unlabeled monocular videos in a self-supervised fashion, resulting in various objective functions \cite{godard2017unsupervised,yin2018geonet,ranjan2019competitive,bian2019unsupervised,godard2019digging,wang2020unsupervised,casser2019depth,wang2022cbwloss,klingner2020self,zhang2020unsupervised,zhang2022self,ruhkamp2021attention,kaushik2021adaadepth,chen2021fixing,tian2021depth}. Such functions are used to seek the global optimal solution for the depth estimation network (DepthNet), and various network architectures \cite{guizilini2020semantically,jung2021fine,zhang2020unsupervised,lyu2021hr,pillai2019superdepth,guizilini20203d,johnston2020self,ruhkamp2021attention,song2021mlda,kaushik2021adaadepth,varma2022transformers,guizilini2022multi} have been designed to build robust DepthNet variants.

However, the existing DepthNet architectures still have many shortcomings. First, to compensate for the loss of fine-grained information caused by plain downsampling operations (e.g., max pooling), the existing methods either ignore (e.g., methods based on plain skip connections \cite{ronneberger2015u}) or fail to exploit (e.g., methods \cite{zhang2020unsupervised,lyu2021hr,zhou2018unet++,huang2020unet} based on both dense connections and lossy downsampling) the fine-grained information contained in lower-level features. To this end, we propose a multilevel near-lossless fine-grained information fusion scheme.
Second, the existing methods fail to adaptively utilize global and local information in parallel during the decoder stage to accurately infer depths. To this end, we propose an adaptive refinement module (AdaRM).
Third, most current methods \cite{bian2019unsupervised,ranjan2019competitive,godard2019digging,wang2020unsupervised,wang2022cbwloss,klingner2020self,guizilini2020semantically,guizilini20203d,lyu2021hr,watson2019self,zhou2017unsupervised} fail to sufficiently and globally analyze the disparity values. To this end, we propose a global disparity attention module (AttDisp).

Specifically, the proposed multilevel near-lossless fine-grained information fusion scheme adaptively incorporates multilevel near-lossless fine-grained information with saliency information, obtained by the proposed adaptive axial-normalized position-embedded channel attention sampling module (AdaAxialNPCAS) from a high-resolution feature map, into a low-resolution feature map with high-level semantics and adaptively fuses the recovered high-frequency information (obtained by constraining the solution space utilizing both the global and local dependencies between pixels) into the high-resolution feature map obtained from the nonlearning method. The proposed AdaRM (a) efficiently captures local information by utilizing a local filter and models long-range dependency-based transformer mechanisms in parallel during the decoding stage, and it (b) adaptively fuses the extracted local and global information into the original feature map. In the proposed AttDisp, we reweight the local disparity map  (regressed from the decoded feature map) by utilizing the global attention weights generated by computing the global correlation of the decoded feature map.

Finally, instead of scaling the prediction results based on median information, to solve the inherent scale ambiguity problem encountered by self-supervised monocular methods \cite{bian2019unsupervised,bian2021unsupervised,godard2019digging,ranjan2019competitive,wang2022cbwloss,wang2020unsupervised,lyu2021hr}, we propose an adaptive scale alignment strategy to scale the obtained estimation results to the ground truths, as measured with light detection and ranging (LiDAR), by considering both median and mean information.

The main contributions of this paper are summarized as follows.
\begin{itemize}
	\item
   We propose a multilevel near-lossless fine-grained information fusion scheme to compensate for the loss of fine-grained information.

	\item
	We propose an adaptive refinement module to efficiently capture local and global information in parallel during the decoding stage and adaptively fuse the extracted local and global information into the original feature map.	
	\item
	We propose a disparity attention module to model the distribution characteristics of disparity values from a global perspective.
	
\end{itemize}

\vspace{-0.2cm}
\section{Related work}\label{sec:relatedwork}

Recently, depth estimation algorithms \cite{zhou2017unsupervised,yin2018geonet,varma2022transformers,lee2022edgeconv,kaushik2021adaadepth,godard2017unsupervised,lee2021patch,guizilini2022multi,ruhkamp2021attention,lyu2021hr,watson2021temporal,wang2020cliffnet,jiao2018look,zhang2020unsupervised,klingner2020self,liu2015learning,ranftl2021vision,parida2021beyond,song2021mlda,fu2018deep,jung2021fine,ranjan2019competitive,guizilini2020semantically,godard2019digging,wang2020unsupervised,zhang2022self,johnston2020self,wang2022cbwloss,garg2016unsupervised,pillai2019superdepth,guizilini20203d,miangoleh2021boosting,bhat2021adabins,yang2021transformer,bian2019unsupervised,eigen2014depth,huynh2020guiding,meng2021cornet,masoumian2022monocular,vyas2022outdoor,gonzalez2021plade,gonzalezbello2020forget,martin2022deconstructing} based on deep learning have attracted considerable attention. These methods can be divided into supervised and self-supervised depth estimation approach depending on whether ground truths are needed. Supervised approaches require expensive depth sensors and considerable labor to obtain sufficient data labeled with pixel-level depth information for network training, while unsupervised methods alleviate this constraint and can directly utilize photometric differences as supervision signals to train neural networks for estimating depths and camera poses from unlabeled monocular videos.

\vspace{-0.3cm}
\subsection{Supervised Depth Estimation}
A supervised depth estimation algorithm requires much manually labeled sparse point cloud data to guide the learning process of the depth network.

Eigen et al. \cite{eigen2014depth} first predicted a depth map from a single image by stacking coarse-scale and fine-scale networks.
Subsequently, the depth estimation task was cast as a deep continuous conditional random field (CRF) learning problem \cite{liu2015learning} or an ordinal regression problem \cite{fu2018deep,meng2021cornet}. 
Jiao et al. \cite{jiao2018look} alleviated the depth data bias issue by modeling depth data statistics based on a depth-aware objective. 
To reduce the sensitivity of the network training process to label noise and outliers, a hierarchical embedding loss was employed as the optimization goal in \cite{wang2020cliffnet}. To further improve the quality of dense depth maps, Miangoleh et al. \cite{miangoleh2021boosting} transferred fine-grained details from a high-resolution input to a low-resolution input by learning an additional conditional generative model.

In the last two years, efforts have been made to leverage attention mechanisms \cite{vaswani2017attention}, in particular, the vision transformer (ViT) \cite{dosovitskiy2020image}, to achieve improved depth estimation. Huynh et al. \cite{huynh2020guiding} exploited nonlocal depth dependencies captured by the depth attention volumes between coplanar points to guide the depth estimation process. Parida et al. \cite{parida2021beyond} modeled the relations between images and echoes and then utilized attention maps to fuse these modalities for depth prediction. Instead of dividing the depth range into a fixed number of intervals \cite{fu2018deep}, Bhat et al. \cite{bhat2021adabins} computed adaptive bins by using a transformer-based architecture block. Instead of transferring the high-frequency details from one estimation to another with structural consistency \cite{miangoleh2021boosting} by learning an additional conditional generative model, Ranftl et al. \cite{ranftl2021vision} leveraged the ViT in place of a convolutional neural network (CNNs) as the encoder to achieve finer-grained and more globally coherent predictions, yielding substantial improvements. Similarly, Yang et al. \cite{yang2021transformer} achieved improved continuous pixelwise prediction by directly combining a linear transformer and ResNet. Different from the above methods \cite{ranftl2021vision,yang2021transformer}, Lee et al. \cite{lee2021patch} modeled the relationships among neighboring pixels using the attention maps of each local patch. Based on \cite{lee2021patch}, the structural information learned from image patches by exploiting graph convolution was also employed in \cite{lee2022edgeconv} to attain improved performance.

The aforementioned depth estimation methods directly cast the depth estimation task as an ordinal regression problem based on the ground truth, process depth bias depending on the ground truth, transfer fine-grained details depending on both the ground truth and an additional conditional generative model, or achieve improved depth prediction based on attention. Although superior performance can obtained by these methods, they heavily depend on the acquired ground-truth data.

\vspace{-0.3cm}
\subsection{Self-Supervised Depth Estimation}
Despite their superior performance, supervised models are not universally applicable and heavily depend on the acquired ground truth-data. Moreover, the data annotation process is often slow and costly. The obtained annotations also suffer from structural artifacts, particularly in the presence of reflective, transparent, and dark surfaces or nonreflective sensors that output infinity values. All these challenges strongly motivate us to infer depth (particularly from monocular videos) in an unsupervised manner.
In 2016,  Garg  et  al. \cite{garg2016unsupervised} first estimated a depth map from a single view in an unsupervised way.
Godard et al. \cite{godard2017unsupervised} performed single-image depth estimation by enforcing left-right depth consistency inside their network, but in these cases, stereo images were required \cite{godard2017unsupervised,garg2016unsupervised}, including the camera motion between a stereo pair \cite{garg2016unsupervised}.
Subsequently, Zhou et al. \cite{zhou2017unsupervised} first proposed a completely unsupervised monocular estimation method.

To explicitly address dynamic objects and occlusions, additional subnetworks \cite{yin2018geonet,ranjan2019competitive,wang2020unsupervised} were introduced to model dynamic objects and static regions, respectively. More geometric prior knowledge \cite{bian2019unsupervised,godard2019digging,wang2022cbwloss} was mined to mitigate the effects of dynamic objects. In addition, semantic information \cite{klingner2020self,guizilini2020semantically,jung2021fine} and multiframe inputs \cite{guizilini2022multi,watson2021temporal} obtained during inference were also used to further achieve improved performance.

Different from the above methods that either optimize objective functions or utilize multitask or mining time information to improve the quality of the resulting depth maps, good DepthNet architectures have also been designed to fit the functions mapped from RGB images to the desired depth levels.

To predict high-quality depth maps, especially depth maps with sharp details, Zhang et al. \cite{zhang2020unsupervised} built dense connections based on DenseNet \cite{huang2017densely} between low-level and high-level features. Similarly, to obtain high-resolution features with spatial and semantic information, Lyu et al. \cite{lyu2021hr} redesigned the skip connections and improved the feature fusion method \cite{zhou2018unet++,hu2018squeeze} utilized between the encoder and decoder. A cross-scale feature fusion scheme \cite{zhao2022monovit,he2022ra,zhou2021self} based on attention was also utilized to infer accurate depths. Pillai et al. \cite{pillai2019superdepth} replaced the naive interpolation operators in their decoder with a subpixel convolutional layer \cite{shi2016real} to capture the fine details in images. Guizilini et al. \cite{guizilini20203d} replaced the standard striding and upsampling operations with packed and unpacked schemes to learn detail-preserving representations. Zhang et al. \cite{zhang2022self} extended the perceptual area of the depth map produced over the source image by utilizing a multiscale scheme.

Similar to supervised models, attention-based schemes have also been adopted to enhance the long-range modeling capabilities of unsupervised depth estimation networks. Johnston et al. \cite{johnston2020self} improved the depth value discretization strategy \cite{fu2018deep} based on self-attention. Ruhkamp et al. \cite{ruhkamp2021attention} explicitly modeled 3D spatial correlations based on self-attention. 
To obtain per-pixel depth maps with sharper boundaries and richer details, Song et al. \cite{song2021mlda} employed a multilevel feature extraction strategy to learn rich hierarchical representations and modeled all the position information contained in the given feature map based on the attention weights calculated for each channel.
Kaushik et al. \cite{kaushik2021adaadepth} forced their model to obtain rich contextual information by directly adding an attention map (computed from the output feature map of the encoder) to the corresponding feature map. Similar to Ranftl et al. \cite{ranftl2021vision}, Varma et al. \cite{varma2022transformers} adopted the ViT as an encoder for self-supervised monocular depth estimation. \added{Han et al. \cite{han2022transdssl} captured long-range dependencies via a transformer backbone during the encoder stage while enhancing the fine details utilizing pixelwise attention during the decoder stage.} Guizilini et al. \cite{guizilini2022multi} utilized attention mechanisms to refine per-pixel matching probabilities, resulting in improvements over the standard similarity metrics, which are prone to ambiguities and local minima.

However, the above methods either ignore or fail to exploit the fine-grained information contained in lower-level features or cannot model global and local information in parallel during the decoding stage. In addition, the existing techniques do not sufficiently and globally analyze the output disparity values. In this paper, we propose (a) a multilevel near-lossless fine-grained information fusion scheme, (b) an adaptive refinement module, and (c) a disparity attention module to address these shortcomings.

\vspace{-0.4cm}
\section{Method}\label{sec:method}

\begin{figure*}[htbp]%[htbp]
	\centering 	
	\includegraphics[scale=0.5]{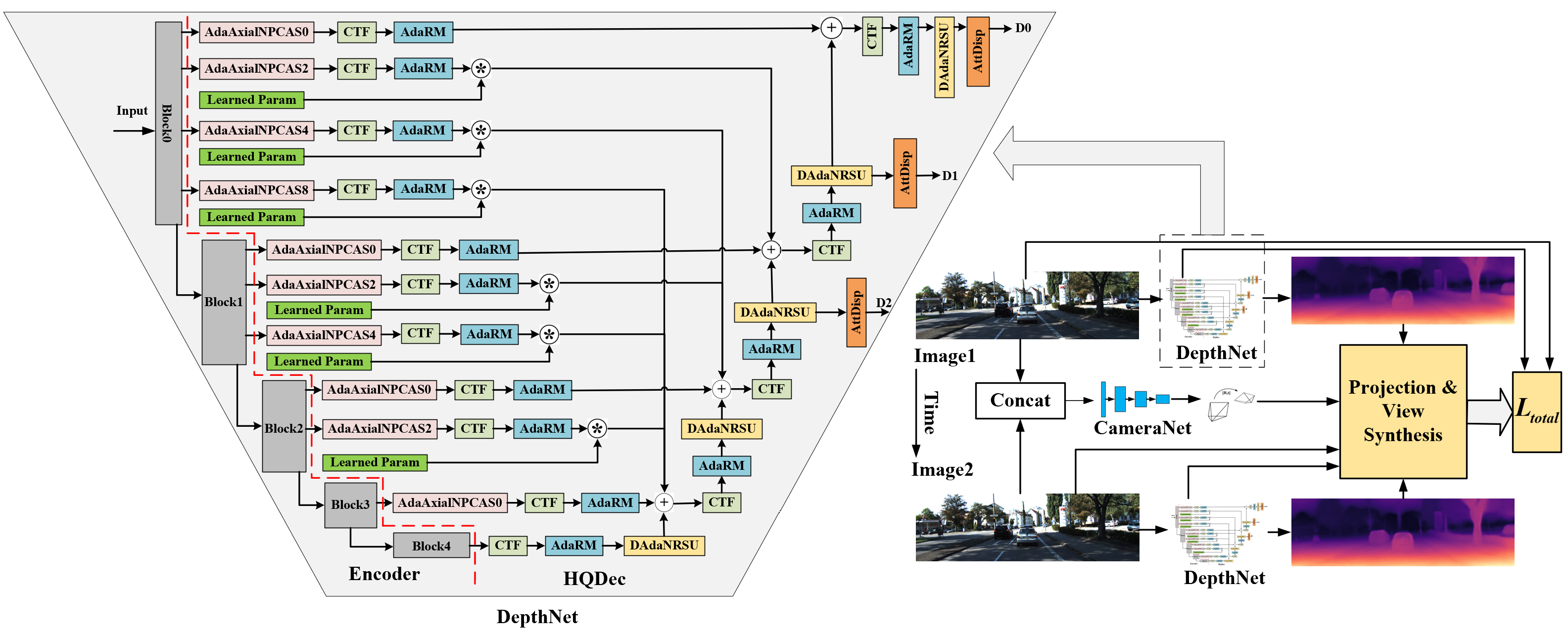}%0.75  scale=0.25
	\caption{Overview. The left part is the diagram of the connection between the encoder and the proposed HQDec. The right panel is a training architecture figure. For simplicity, we only draw the one-way training process here, and the other direction only requires the roles to be switched, except when calculating the inverse pose. `Block' represents the block module generated by dividing the backbone according to the resolution of the output feature map. `Learned Param' represents an elementwise learnable parameter with an initial value of zero, which is used to control the propagation of information. `CTF' represents the channel transformation function. `AdaRM' represents the adaptive refinement module proposed in Section \ref{subsec:feat_refine}. `AdaAxialNPCAS' represents the downsampling module proposed in Section \ref{channel_attention_sample}. `DAdaNRSU' represents the upsampling module proposed in Section \ref{upsample_module}. `AttDisp' represents the disparity attention module proposed in Section \ref{sec:global_disp_att_module}.}\label{fig:overview}	
	\vspace{-10pt}
\end{figure*}

The overview of the proposed high-quality decoder (HQDec) is shown in Fig. \ref{fig:overview}.

\vspace{-0.2cm}
\subsection{Problem Description and Optimization Objective}
In the self-supervised monocular depth estimation task, our goal is to learn a function {\small $D_{tgt}=F_{D}(I_{tgt}|W_{D})$} that can infer the corresponding scene depth {\small $D_{tgt}$} from the given image {\small $I_{tgt}$}, where {\small $W_{D}$} is the learned weight. Due to the lack of ground-truth depth values, which are used to guide the network parameter updates, we need a subnetwork to learn a function {\small $T=F_{T}([I_{ref},I_{tgt}]|W_{T})$} that can be used to predict the relative pose between the reference image and the target image. Following a previously developed method \cite{wang2022cbwloss}, the loss function shown in formula \eqref{eq_totalloss} is used for the optimization objective to jointly train DepthNet and CameraNet on unlabeled monocular videos.
\begin{equation}
\begin{aligned}
\label{eq_totalloss}
\added{	L_{total}=\lambda_{p}*\hat L_{p}+\lambda_{d}*L_{d}+\lambda_{f}*L_{feat}+L_{s}}
\end{aligned}
\end{equation}
where the hyperparameter settings in formula \eqref{eq_totalloss} are the same as those in the previous work \cite{wang2022cbwloss}. $\hat L_{p}$ represents the bidirectional weighted photometric loss proposed in \cite{wang2022cbwloss}, $L_{d}$ represents the bidirectional depth structure consistency loss proposed in \cite{wang2022cbwloss}, and $L_{s}$ represents the smoothness loss used in \cite{wang2022cbwloss}. $L_{feat}$ represents the bidirectional weighted feature perception loss employed by the bidirectional feature perception loss proposed in \cite{wang2022cbwloss}, which is weighted by both the adaptive weights and the bidirectional camera occlusion masks proposed in \cite{wang2022cbwloss}.

\vspace{-0.3cm}
\subsection{AdaRM}\label{subsec:feat_refine}
\begin{figure}[htbp]%[htbp]
	\vspace{-8pt}
	\centering 	
	\includegraphics[scale=0.38]{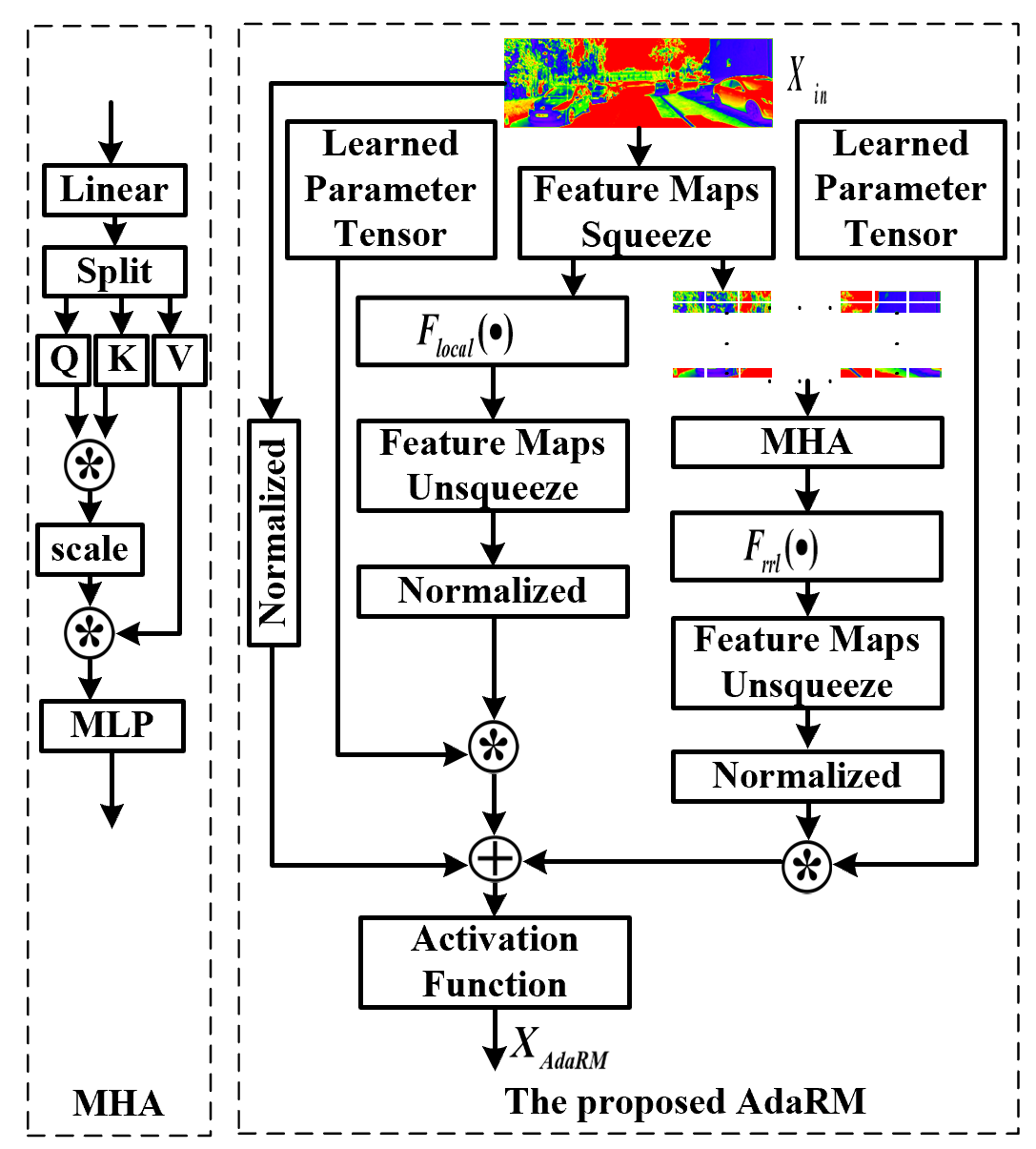}	%0.38
	\caption{\replaced{The proposed feature map refinement module.}{The two proposed schemes for feature map refinement.}}\label{fig:fine_tuned}	
	%\vspace{-10pt}
\end{figure}

It has been the consensus that the local feature information in an image can be efficiently extracted by a CNN, but such a network cannot model the long-range dependencies between pixels. Although the receptive field of the feature map can be increased by downsampling, which can be used to efficiently perform image classification tasks, the fine-grained information contained in a high-resolution feature map is gradually discarded as multilevel downsampling is performed. It is difficult to recover the lost information from a low-resolution feature map, which is not conducive to dense prediction tasks. On the other hand, although pure transformer mechanisms \cite{vaswani2017attention,dosovitskiy2020image} have unique advantages in terms of modeling global dependencies, they lack some of the inductive biases that are inherent to CNNs, such as translation equivariance and locality. Although the use of this pure transformer mechanism can be efficient for image classification tasks, which depend on the use of more global information to distinguish the differences between objects, and machine translation tasks, which rely on the context information of words or sentences, such a mechanism is not also conducive to dense prediction tasks, which need to account for both the local pixel context information and the global information of each object in the given image.
This is different from previously developed approaches \cite{vaswani2017attention,dosovitskiy2020image}, which draw long-term dependencies between sequences of input words \cite{vaswani2017attention} for machine translation tasks or sequences of linear patch embeddings split from images \cite{dosovitskiy2020image} for image classification tasks by utilizing pure multihead self-attention (MHA) mechanisms. To this end, we exploit the strengths of both CNNs and MHA and propose an AdaRM, as shown in Fig. \ref{fig:fine_tuned}. Concretely, the information of $X_{AdaRM}$ comes from the sum of the information of the three subbranches. The information in the left branch is the original information contained in $X_{in}$. This branch ensures that the information in $X_{in}$ can be fully propagated to the next stage ($X_{AdaRM}$). In the middle branch, we utilize CNNs to further model the local context information of the pixels in $X_{in}$ and adaptively propagate it to $X_{AdaRM}$ by employing a learnable parameter tensor, which guarantees that the information of each pixel in $X_{AdaRM}$ is obtained by performing weighted fusion on the pixel information at the corresponding position and the surrounding positions in $X_{in}$. In the right branch, we utilize MHA to draw global dependencies between the sequences of subfeatures split from $X_{in}$ and adaptively propagate them to $X_{AdaRM}$ by employing a learnable parameter tensor, which ensures that the information in each subfeature map depends on the information in all remaining subfeature maps. As a result, each pixel in $X_{AdaRM}$ has the ability to perceive both local and global information without downsampling.

Specifically, suppose we are a feature map $X \in \Re^{C\times H\times W}$, where $C$, $H$, and $W$ are the number of channels, height, and width of the feature map, respectively. A feature map $X_{3}$ with local context information can be obtained by the function $F_{local}(\cdot)$ with a square filter (e.g., $3\times3$) according to formula \eqref{eq_local}. To make the information denser, we first squeeze $X$ by employing the $F_{squeeze}(\cdot )$ function (e.g., a $1\times 1$ filter) before extracting local context information and then unsqueeze $X_2$ to its original dimensionality by utilizing $F_{unsqueeze}(\cdot)$ (e.g., a $1\times 1$ filter) for the subsequent feature fusion process.
\vspace{-0.1cm}
\begin{subequations}
	\label{eq_local}
	\begin{equation}\label{eq_local2}	
	X_{1}=F_{squeeze}(X),X_1\in \Re^{\frac{C}{N_{sq}}\times H \times W}	
	\end{equation}
	\begin{equation}	
	X_2=F_{local}(X_{1}),X_2\in \Re^{\frac{C}{N_{sq}}\times H \times W}	
	\end{equation}
	\begin{equation}	
	X_3=F_{unsqueeze}(X_2),X_3\in \Re ^{C\times H\times W}	
	\end{equation}	   
\end{subequations}
where $N_{sq}$ represents the squeezing ratio.

\begin{table}[htbp] \small% 
	\vspace{-2pt}	
	\setlength\tabcolsep{2pt}
	\centering	
	\renewcommand\arraystretch{1.5}	
	\begin{tabular}{lcccccccccccccccc} %需要10列
		\toprule 
		\multicolumn{1}{c}{\multirow{1}*{}}&\multicolumn{5}{c}{\multirow{1}*{Fine-tume Module}}& & & &\multicolumn{3}{c}{\multirow{1}*{Disp Module}}&\\
		\multicolumn{1}{l}{}&\multicolumn{1}{c}{$s_0$}&\multicolumn{1}{c}{$s_1$} &  \multicolumn{1}{l}{$s_2$} &\multicolumn{1}{c}{$s_3$}& \multicolumn{1}{c}{$s_4$}& & & & \multicolumn{1}{c}{$s_0$}& \multicolumn{1}{c}{$s_1$}& \multicolumn{1}{c}{$s_2$} \\
		
		\hline %		
		
		\multicolumn{1}{l}{$N_{sq}$}&2&4&4&8&8& & & &-&-&-\\	
		\hline
		\multicolumn{1}{l}{$H_{sub}$}&$\frac{H_{in}}{16}$&$\frac{H_{in}}{16}$&$\frac{H_{in}}{16}$&$\frac{H_{in}}{16}$&$\frac{H_{in}}{32}$ & & & &$\frac{H_{in}}{16}$&$\frac{H_{in}}{16}$&$\frac{H_{in}}{16}$ \\
		
		\hline
		
		\multicolumn{1}{l}{$W_{sub}$}& $\frac{W_{in}}{16}$&$\frac{W_{in}}{16}$&$\frac{W_{in}}{16}$& $\frac{W_{in}}{16}$&$\frac{W_{in}}{32}$& & & &$\frac{W_{in}}{16}$&$\frac{W_{in}}{16}$&$\frac{W_{in}}{16}$\\
		
		\hline
		\multicolumn{1}{l}{$C_{embed}$}&24&48&64&160&256& & & &12&24&32\\		
		
		\hline	
		\multicolumn{1}{l}{$N$}&2&4&4&8&8& & & &2&4&4\\

		\bottomrule 
	\end{tabular}
	\caption{Parameter settings.}\label{tab:hwc_param}
	 \vspace{-10pt}
\end{table}

To obtain a feature map with global context information, we sample $X_1$ into a nonoverlapping subfeature map by employing the sample function $F_{sample}(\cdot)$, which can be implemented with a filter whose shape is the same as that of the subfeature map and whose cross-correlation stride is controlled according to the size of the filter. The subfeature map can be embedded into a feature vector $x$, which is employed to calculate the global correlation of the subfeature map via a learnable embedding function $F_{embed}(\cdot)$, as shown in formula \eqref{eq_embed}. \vspace{-0.1cm}
\begin{equation}
\begin{aligned}
\label{eq_embed}
x=F_{embed}(F_{sample}(X_1)),x\in \Re^{\frac{H}{H_{sub}}*\frac{W}{W_{sub}}\times C_{embed}}
\end{aligned}
\end{equation}
where $H_{sub}$, $W_{sub}$, and $C_{embed}$ denote the height and width of the subfeature map and the number of feature vectors, respectively. In $F_{embed}(\cdot)$, we first utilize a $1\times 1$ filter followed by an ELU action function to embed the sampled feature map into the desired dimension $C_{embed}$, flatten the embedded feature map into a one-dimensional tensor, and finally transpose it into the desired feature vector x.

To compute the global correlation weights of $x$, we first triple $x$ by utilizing a learnable linear transformation function $F_{linear}(\cdot)$ (e.g., a linear layer), and then we reshape these feature vectors to $N$ subspaces via the function $F_{reshape}(\cdot)$ for the purpose of jointly focusing on the information derived from different representation subspaces at different positions. Thereafter, the feature vectors of each subspace are split into three parts, as shown in formula \eqref{eq_qkv1}; one of these parts serves as the query vector, and the other two parts serve as the key-value pairs. We can compute the global correlation weights by rescaling the product of the query vector and the corresponding key via the softmax function in each subspace. These weights are assigned to the corresponding value vectors, resulting in corresponding global feature vectors in the corresponding subspace. Then, the global feature vector $x_1$ containing the information from different representation subspaces at different positions can be obtained by concatenating and linearly transforming different subfeatures in different subspaces. Note that we adopt residual connections\cite{he2016deep} to mitigate the degradation problem and utilize a multilayer perception function $F_{mlp}(\cdot)$, implemented by nonlinear transformation layers consisting of two linear layers followed by a Gaussian error linear unit function, to enhance the nonlinear fitting ability of the model. The final global feature vector $x_2$ in formula \eqref{eq_qkvx2} can then be obtained.

\vspace{-0.1cm}
\begin{subequations}
	\label{eq_qkv}
	\begin{equation}
	\begin{aligned}
	\label{eq_qkv1}
	Q,K,V=F_{split}(F_{reshape}(F_{linear}(x)))
	\end{aligned}
	\end{equation}
	\begin{equation}
	\begin{aligned}
	\label{eq_qkvx}
	x_1=F_{linear}(F_{reshape}(Softmax(QK^{T}*s)V))
	\end{aligned}
	\end{equation}
	\begin{equation}
	\begin{aligned}
	\label{eq_qkvx2}
	x_2=x+x_1+F_{mlp}(x+x_1)
	\end{aligned}
	\end{equation} 	
\end{subequations}
where $Q,K,V \in \Re^{N\times \frac{H}{H_{sub}}*\frac{W}{W_{sub}}\times \frac{C_{embed}}{N}}$, $x_1\in \Re^{\frac{H}{H_{sub}}*\frac{W}{W_{sub}}\times C_{embed}}$,  $x_2\in \Re^{\frac{H}{H_{sub}}*\frac{W}{W_{sub}}\times C_{embed}}$. $s=(\frac{C_{embed}}{N})^{-0.5}$ denotes the scale factor.

For information fusion purposes, the dimensions of $x_2$ and $X$ must be equal. To this end, $x_2$ is first mapped to subfeature map $X_4$ in formula \eqref{eq_x4} by employing the learned function $F_{rrl}(\cdot)$, which can be implemented by a transpose convolution whose filter shape and stride are the same as those of the subfeature map. Then, the channels of $X_4$ in formula \eqref{eq_x5} are aligned by a learned linear function $F_{unsqueeze}(\cdot)$ (e.g., a $1\times 1$ filter). Finally, $X_{AdaRM}$ is obtained according to formula \eqref{eq_ada_rm}.

\vspace{-0.1cm}
 \begin{subequations}
 	\label{eq_refine}
 	\begin{equation}
 	\begin{aligned}
 	\label{eq_x4}
 	X_4=F_{rrl}(x_2),X_4\in \Re ^{\frac{C}{N_{sq}}\times H \times W}
 	\end{aligned}
 	\end{equation}
 	\begin{equation}
 	\begin{aligned}
 	\label{eq_x5}
 	X_5=F_{unsqueeze}(X_4),X_5\in \Re^{C\times H \times W}
 	\end{aligned}
 	\end{equation} 		
 	\begin{equation}
 	\begin{aligned}
 	\label{eq_ada_rm}
 	{\footnotesize {\tiny X_{{\tiny AdaRM}}=X+P_{1}*X_{3}+P_{2}*X_{5},	X_{{\tiny AdaRM}}\in  \Re^{C\times H \times W}}}
 	\end{aligned}
 	\end{equation}
 \end{subequations} 
where $P_1,P_2 \in \Re^{H \times W}$ denotes learnable parameter tensors with initial values of zero.

\vspace{-0.2cm}

\vspace{-0.1cm}

\subsection{Multilevel Near-Lossless Fine-Grained Information Fusion}\label{infor_exchange}
Fine-grained information has a significant impact on the performance achieved in dense prediction tasks. However, the loss of high-frequency information occurs as downsampling proceeds. Once high-frequency information is lost, it is difficult to recover fine-grained high-resolution feature maps from low-resolution maps after performing noninvertible low-pass filtering and subsampling operations \cite{wang2020deep,shi2016real}.
The existing methods design either plain skip connections \cite{ronneberger2015u} or dense connections \cite{huang2020unet} between the encoder and decoder to reduce the degree of information loss. However, these schemes either ignore or fail to exploit the fine-grained information contained in lower-level features. To this end, we propose a multilevel near-lossless fine-grained information fusion scheme to compensate for the loss of fine-grained information. In the proposed scheme, we address this challenge by (1) improving the downsampling strategy to propagate more fine-grained information to low-resolution feature maps, (2) adaptively incorporating multilevel fine-grained information into a high-level feature map, and (3) improving the upsampling strategy to recover as much high-frequency information as possible.

\subsubsection{Drive the low-resolution feature map to retain more fine-grained information}\label{channel_attention_sample}
\begin{figure}[htbp]%[htbp]
	\centering 	
	\includegraphics[height=85mm,width=50mm]{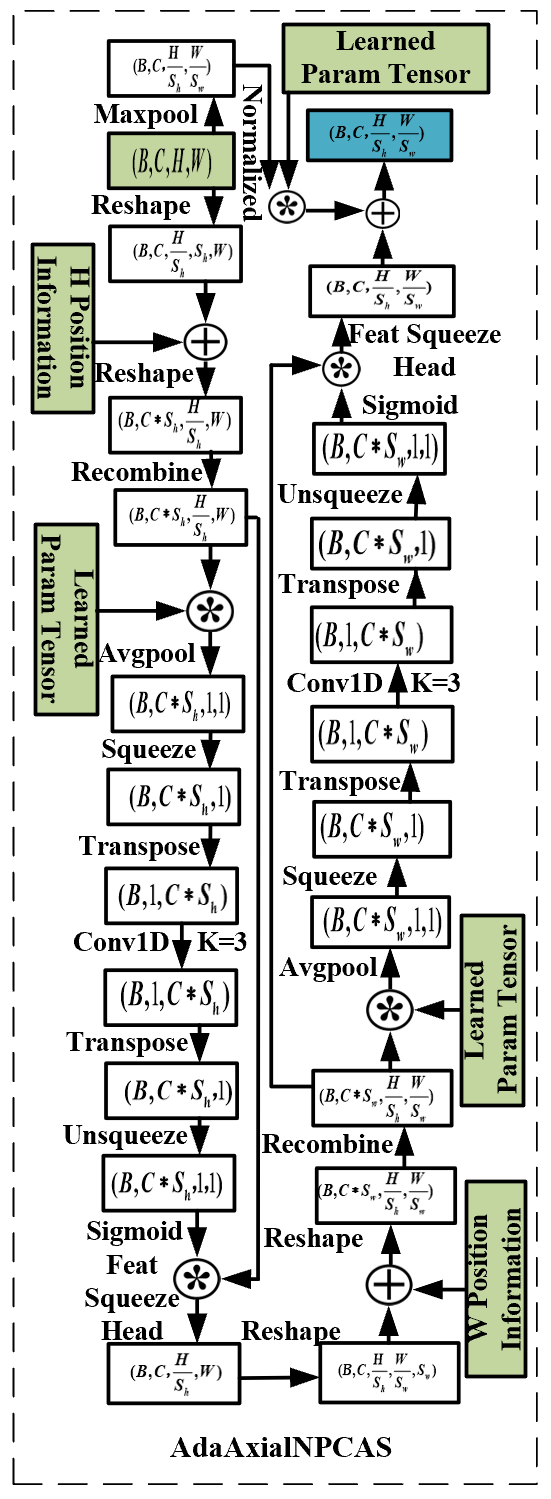}
	\caption{The proposed downsampling schemes.}\label{fig:downsample}	
	\vspace{-10pt}
\end{figure}

Compared to the low-resolution feature map with high-level semantics output by the encoder, the feature map with a higher resolution output in the shallow layer preserves more details. This motivates us to propagate more of the spatial structure information contained in the high-resolution feature map to the low-resolution features with high-level semantics. Inspired by \cite{shi2016real,guizilini20203d,hu2018squeeze,woo2018cbam}, we propose AdaAxialNPCAS, as shown in Fig. \ref{fig:downsample}. Instead of directly performing max pooling or stride downsampling, the proposed AdaAxialNPCAS directly folds the spatial dimensions of convolutional feature maps into extra feature channels along the axial direction to obtain the corresponding low-resolution feature maps. The fact that the extra pixels in high-resolution feature maps are directly put into extra channels in low-resolution feature maps during transformation ensures that information is not lost. Different from a previously developed approach \cite{guizilini20203d}, which compresses the concatenated feature maps to a desired number of output channels via 3D convolutions, we first add the position information of each element in the original feature map before executing recombination, and then compute a channel head that can represent each channel by utilizing learnable global average pooling. To give greater weights to the required channels, we reweight each channel by employing the weights calculated by the channel head and compress the reweighted feature maps to a desired shape.

Concretely, given a feature map $X \in \Re^{C\times H \times W}$, we first rearrange the elements in $X$  along the height direction and add the position information of each element in the original feature map to the rearranged tensor for the purpose of losslessly squeezing the spatial information into the channel dimension, resulting in a new tensor $X_{h} \in \Re^{C*S_{h} \times \frac{H}{S_{h}} \times W}$, where $S_{h}$ denotes the number of downsampling operations along the height dimension. Then, we recombine the information contained in $X_h$ by utilizing group convolution. To model the dependencies between the channels of the recombined feature map, we first weight the elements in the feature map by utilizing a learned parameter tensor, which is initialized to one, because different elements in the same channel play different roles in representing global channel information. We then perform global average pooling on the spatial information of the weighted feature map to represent each channel and employ a 1D convolution with a kernel of size 3 to learn correlations. Finally, the attention weights between the different channels can be obtained with a sigmoid function. The recombined feature map, whose channels contain the spatial information of the original feature map, is multiplied by the attention weights to let the model focus on the more useful spatial information generated by squeezing the spatial structure of the original feature map into the channel dimension. We then squeeze the channel to the desired shape by utilizing a feature map squeezing head (e.g., $3\times 3$ convolutions followed by an ELU function). Similarly, we process the obtained feature map along the width direction. To preserve the most salient information contained in the low-resolution feature map and adaptively fuse it into the feature map obtained by the above processing scheme, the sampled feature map with the salient information, obtained by max pooling, is normalized and multiplied by a learned parameter tensor, which is initialized to zero, before being fused.

\vspace{-0.3cm}

 \subsubsection{Adaptively incorporate multilevel fine-grained information into a high-level feature map}\label{adaptive_param_module}

Among the existing works \cite{huang2020unet,lyu2021hr}, reference \cite{huang2020unet} directly fused  lower-level detailed information with high-level semantic feature maps; however, the existing semantic gaps between different levels of information impede information fusion. Reference \cite{lyu2021hr} fused different-level semantic information into higher-resolution feature maps. However, inaccurate semantic information has a negative impact on higher-resolution feature maps.

To recover scene depth as accurately as possible by utilizing multilevel hierarchical information, we propose AdaIE, which can adaptively incorporate multilevel fine-grained information into a high-level semantic feature map and let the model itself decide what fine-grained information needs to be fused and to what extent. Concretely, to control the propagation of information, we first create a learnable parameter tensor that has the same shape as the feature map to be fused and set its initial value to zero. Before the spatial structure details contained in the high-resolution feature map are incorporated into the high-level semantic feature map, the tensor is multiplied by the corresponding feature map, as shown in formula \eqref{eq_infor_exchange}. For example, the encoded feature map $X_{enc}^{3}$ is not only embedded with higher-level semantic information from $X_{dec}^{4}$ but also adaptively fused with the spatial structure details contained in the lower-level feature maps (e.g., $X_{enc}^0$, $X_{enc}^{1}$, etc.) before being decoded. Consequently, higher-level semantic information and more accurate spatial details can achieve complementary advantages, resulting in sharper scene depths.

\vspace{-0.1cm}
\begin{equation}
\begin{aligned}
\label{eq_infor_exchange}
X_{dec}^{k} \sim \begin{cases}
X_{dec}^{k+1}+X_{enc}^{k} ,k=0 \\
X_{dec}^{k+1}+X_{enc}^{k}+\sum _{i=0}^{k-1}P^iX_{enc}^{i}, k>0
\end{cases}
\end{aligned}
\end{equation}
where $P^i$ is a learnable parameter tensor whose initial value is set to zero. Its parameter is updated along with the model parameters during training. $k \in \{0,1,2,3\}$ represents the number of stages.

 \subsubsection{Recover as much high-frequency information as possible by constraining the solution space}\label{upsample_module}
\begin{figure}[htbp]%[htbp]
	\centering 	
	%\vspace{-10pt}
	\includegraphics[scale=0.45]{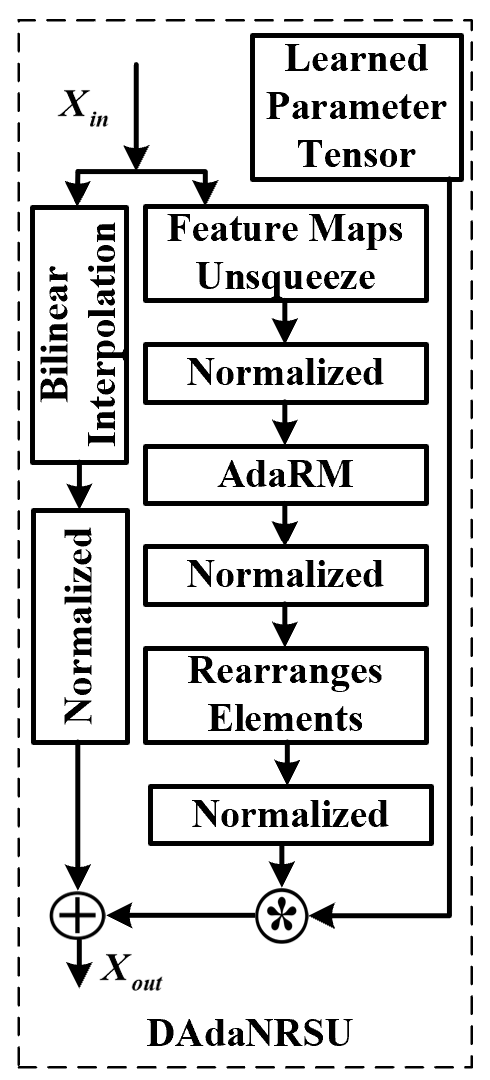}%0.45   0.18
	\caption{The proposed upsampling schemes.}\label{fig:fine_upsample}	
	\vspace{-10pt}
\end{figure}

Upsampling, which directly affects the quality of the predicted depth maps, is an important component of DepthNet that is based on an encoder and a decoder. However, it is difficult to recover fine-grained high-resolution feature maps from low-resolution maps by utilizing nonlearning methods because the process of mapping from a low-resolution feature space to a high-resolution feature space is a one-to-many problem that may have multiple solutions. To this end, different from the existing methods that recover high-resolution feature maps with only bilinear or nearest interpolation, in which only the local context information can be considered, we propose DAdaNRSU (shown in Fig. \ref{fig:fine_upsample}), which recovers more high-frequency information by adaptively modeling the local and global dependencies between pixels to restrict the solution space and adaptively fuse this information into the coarse-grained high-resolution feature maps obtained by traditional upsampling methods.

\vspace{-0.2cm}

Specifically, given a feature map $X_{in} \in \Re^{C\times H\times W}$, on the one hand, the coarse-grained high-resolution feature map $X_{high}^1\in \Re^{C\times 2H\times 2W}$, which helps with providing a good initial value for each layer of the decoder at the beginning of the training process, can be recovered via traditional upsampling methods, which are commonly used in decoders \cite{zhou2017unsupervised,bian2019unsupervised,ranjan2019competitive}. On the other hand, the low-resolution feature map $X_{in}$ can be expanded by a learned function $F_{expand}(\cdot)$ with a $1\times 1$ filter into a higher-dimensional feature subspace to recover the high-resolution feature map in formula \eqref{eq_refine_upsample}. To seek additional constraints that limit the solution space, we model the local and long-range dependencies between the pixels in $X_{low}^1$ by utilizing formula \eqref{eq_ada_rm} described in Section \ref{subsec:feat_refine}. Then, the high-resolution feature map $X_{high}^2 \in \Re^{C \times 2H\times 2W}$ can be obtained by rearranging the elements in $X_{low}^2 \in \Re^{4C \times H \times W}$ to form a new feature space  $\Re ^{C \times 2H \times 2W}$ by employing the pixel shuffling function  $F_{pixelshuffle}(\cdot)$ from \cite{paszke2017automatic}. Finally, we adaptively fuse  $X_{high}^2$, normalized by formula \eqref{norm_eq}, into a normalized version $X_{high}^1$ by utilizing a learned parameter tensor $P$ that has the same shape as the feature map and is initialized to zero.

%\vspace{-0.2cm} 
\begin{subequations}
	\label{eq_refine_upsample}
	\begin{equation}
	\begin{aligned}
	X_{low}^ {1}=F_{expand}(X_{in}), X_{low}^1 \in \Re^{4C \times H \times W}
	\end{aligned}
	\end{equation}
	\begin{equation}
	\begin{aligned}		
	X_{low}^{2}=F_{refine}(X_{low}^{1}),X_{low}^{2}\in \Re^{4C\times H \times W}
	\end{aligned}
	\end{equation}	
	\begin{equation}
	\begin{aligned}
	\label{eq_refine_upsample_pixelshuffle}
	X_{high}^2=F_{pixelshuffle}(X_{low}^{2}),X_{high}^2\in \Re^{C \times 2H\times 2W}
	\end{aligned}
	\end{equation}
	\begin{equation}
	\begin{aligned}	
	X_{high}=X_{high}^{1}+P*X_{high}^{2}, ,X_{high} \in \Re^{C\times 2H \times 2W}
	\end{aligned}
	\end{equation} 
	   \begin{equation}\label{norm_eq}
	   \begin{aligned}
	   Y=\frac{X-E(X)}{\sigma(X)}
	   \end{aligned}
	   \end{equation}
\end{subequations}
where $E(\cdot)$ and $\sigma(\cdot)$ are the mean standard deviation, respectively.

\vspace{-0.4cm}
\subsection{Disparity Attention Module}\label{sec:global_disp_att_module}
\begin{figure}[htbp]%[htbp]
	\centering 
	\vspace{-5pt}	
	\includegraphics[scale=0.13]{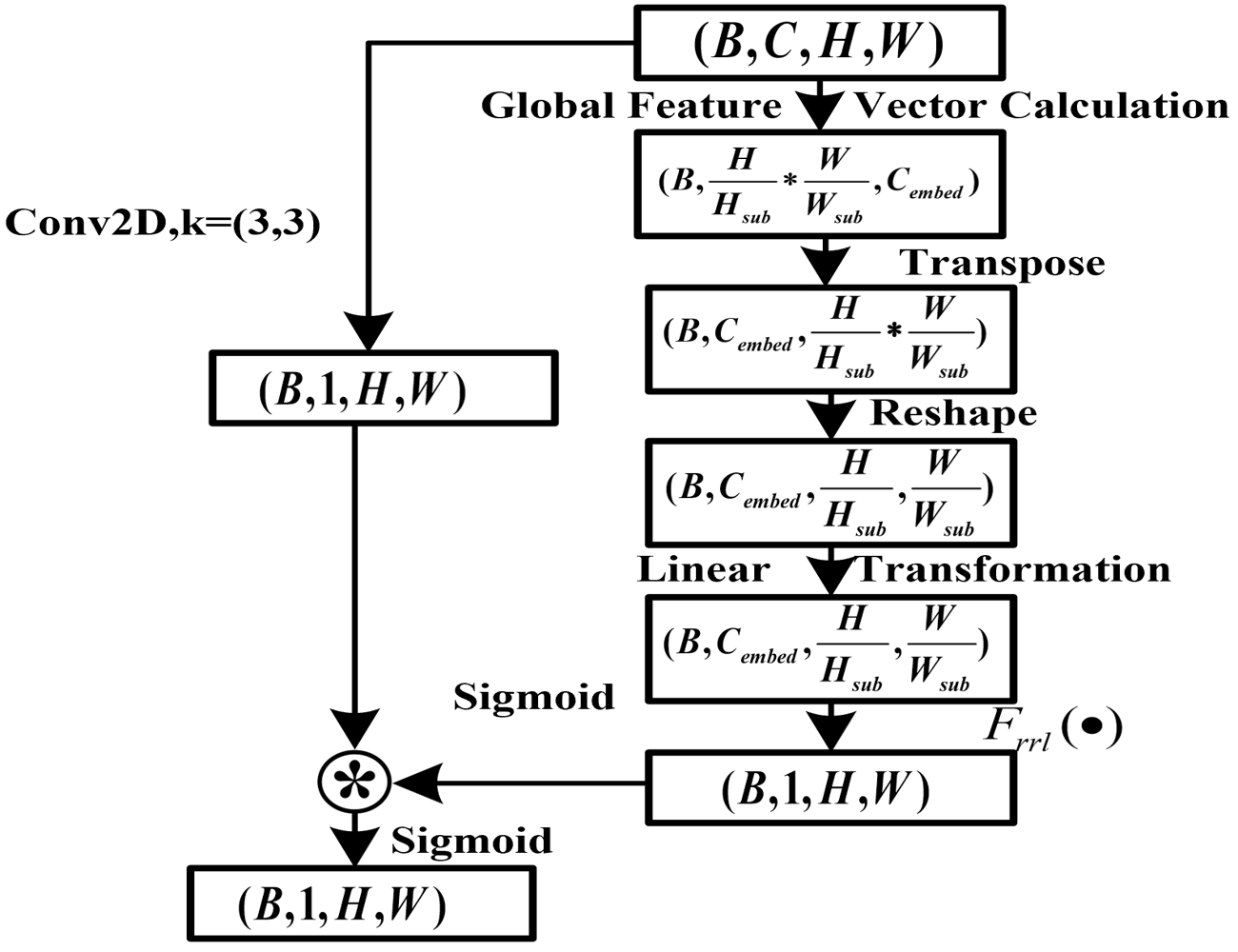}%0.4
	\caption{The proposed disparity attention module.}\label{fig:disp_attention}	
	\vspace{-15pt}
\end{figure}

The disparity output layer, which plays an important role in the process of transforming the decoded feature map into the desired disparity information, is an important part of the depth estimation network. However, most current methods \cite{bian2019unsupervised,ranjan2019competitive,godard2019digging,wang2020unsupervised,wang2022cbwloss,klingner2020self,guizilini2020semantically,guizilini20203d,lyu2021hr,watson2019self,zhou2017unsupervised} directly use local 2D convolution followed by a sigmoid function to regress the decoded feature map to the disparity value, but this technique is incapable of performing a sufficient global analysis for inferring the disparity value of the current pixel. Intuitively, sufficient context information can provide better semantic pixel information and help accurately predict the depth information of the current pixel. To this end, we propose a disparity attention module (shown in Fig. \ref{fig:disp_attention}) to infer the disparity value from the corresponding decoded feature map by utilizing sufficient context information.

Specifically, given the decoded feature map $X_{dec} \in \Re^{C\times H \times W}$, we can compute the corresponding global feature vectors $x_{dec} \in \Re^{\frac{H}{H_{sub}}*\frac{W}{W_{sub}} \times   C_{embed}}$ according to formulas \eqref{eq_embed} and \eqref{eq_qkv}. Different from the purpose of the global feature vectors in Section \ref{subsec:feat_refine}, we expect to obtain the global correlation matrix of the disparity information from $x_{dec}$. To this end, we first reshape $x_{dec}^T \in \Re^{C_{embed} \times \frac{H}{H_{sub}}*\frac{W}{W_{sub}}}$ into a matrix and then utilize learnable linear functions (e.g., 1$\times$1 convolution functions) to linearly transform this matrix into the corresponding subfeature map $X_{gc} \in \Re ^{C_{embed} \times \frac{H}{H_{sub}} \times \frac{W}{W_{sub}}}$. We then map $X_{gc}$ into the attention feature map  $X_{att} \in \Re ^{{1} \times H \times W}$ by utilizing the learned function $\hat F_{rrl}(\cdot)$ implemented by a transpose convolution, whose filter shape and stride are the same as those of the subfeature map. In parallel, we map $X_{dec}$ into the corresponding disparity map $D_{local} \in \Re ^{1\times H \times W}$ by utilizing a 2D convolution kernel. To endow 
$D_{local}$ with a global visual field, $D_{local}$ is multiplied by the global attention weights generated by activating $X_{att}$ using a sigmoid function.

\vspace{-0.5cm}
\subsection{Adaptive Scale Alignment Strategy}\label{sec:scale_alignment_strategy}
The depth scales predicted directly from monocular videos are unknown. Instead of calculating the scale factor based on median information \cite{bian2019unsupervised,bian2021unsupervised,godard2019digging,ranjan2019competitive,wang2022cbwloss,wang2020unsupervised,lyu2021hr}, we propose an adaptive scale alignment strategy, shown in formula \eqref{eq_ada_fuse}, where both the median and mean information can be considered to obtain the scale factor.

\vspace{-0.15cm}

\begin{equation}
\begin{aligned}
\label{eq_ada_fuse}
scale_{ada}=\frac{\zeta * D_{median}^{gt}+(1-\zeta )*D_{mean}^{gt}}{\zeta * D_{median}^{pred}+(1-\zeta )* D_{mean}^{pred}}
\end{aligned}
\end{equation}

The median scale is a special case in which $\zeta =1$. We divide the interval from 0 to 1 into 10 equal pieces. For each value, we calculate the corresponding relative absolute value error, and the scale factor corresponding to the minimum error is used as the scale factor of the current frame depth. 

\vspace{-0.3cm}
\subsection{Network Design}\label{sec:network_design}

\paragraph{DepthNet}
EfficientNetV2-s \cite{tan2021efficientnetv2}, without a classifier, is used as an encoder. To build DepthNet, the encoder is divided into five blocks according to the resolution of the output feature map. The encoded feature map output by each block is refined according to formula \eqref{eq_refine_down}. The overall connection diagram between the encoder and the HQDec is shown in Fig. \ref{fig:overview}.

\vspace{-0.2cm}
\begin{equation}
	\begin{aligned}
		\label{eq_refine_down}
		X_{enc}^{(i,j)}=F_{refine}(F_{ct}(F_{down}^{(i,j)}(X_{enc}^i)))
	\end{aligned}
\end{equation}
where $i\in \{0,...,3\},j\in \{0,...,3-i\}$, $F_{down}^{(i,j)}(\cdot)$ denotes the downsampling module proposed in Section \ref{channel_attention_sample}, and $F_{down}^{(i,j)}(X_{enc}^i)$ means that the encoded feature map $X_{enc}^i$ at stage $i$ is downsampled $2^{j}$ times by employing the CAS or its variant proposed in Section \ref{channel_attention_sample}. $F_{ct}(\cdot)$ represents the channel transformation function that is implemented by a $3\times 3$ filter followed by an ELU function. $F_{refine}(\cdot)$ denotes the feature map refinement module proposed in Section \ref{subsec:feat_refine}.

Therefore, the decoded feature map $X_{dec}^{k}$ at stage $k$ can be obtained by rewriting formula \eqref{eq_infor_exchange}, and the calculation process is shown in formula \eqref{eq_dec_refine}.
\vspace{-0.2cm}
\begin{equation}
\begin{aligned} 
\label{eq_dec_refine}	
X_{dec}^k=\begin{cases} F_{up}(F_{refine}(X_{enc}^4))+X_{enc}^{(3,0)}\\ +\sum\limits_{n=0}^{2}P^{(i,3-n)}X_{enc}^{(i,3-n)}, \qquad k=3 \\
F_{up}(F_{refine}(X_{dec}^{k+1})) +X_{enc}^{(k,0)}\\+\sum\limits_{n=0}^{k-1}P^{(i,k-n)}X_{enc}^{(i,k-n)},\qquad 0 < k \le 2  \\ F_{up}(F_{refine}(X_{dec}^{1})) +X_{enc}^{(0,0)}, \qquad k=0 \end{cases}
\end{aligned}
\end{equation}

where $F_{up}(\cdot)$ denotes the fine-tuning and upsampling function proposed in Section \ref{upsample_module}, and $P$ represents a learnable parameter tensor that has the same shape as the corresponding feature map in Section \ref{adaptive_param_module}.

Following previous work\cite{wang2022cbwloss}, the decoded feature map $X_{dec}^{k},k \in \{0,1,2 \}$ is used as the corresponding candidate feature to generate disparity information. According to formula \eqref{disp_func}, we map the decoded feature map $X_{dec}^{k}$ to the desired disparity. Finally, as is common practice \cite{wang2022cbwloss,ranjan2019competitive,bian2019unsupervised,zhou2017unsupervised}, the estimated disparity value is mapped into an actual distance between $0.1$ and $100$ meters according to formula \eqref{eq_D}.
\vspace{-0.2cm}
\begin{equation}
\begin{aligned}\label{disp_func}
D_{disp}^{k}=F_{disp}(F_{up}(F_{refine}(F_{ct}(X_{dec}^{k}))))
\end{aligned}
\end{equation}
\begin{equation}
\begin{aligned}
\label{eq_D}	
%\hat D^k=\frac{1}{10*D_{disp}^{k}+0.01}
\hat D^k=1/(10*D_{disp}^{k}+0.01)
\end{aligned}
\end{equation}
where $F_{disp}(\cdot)$ represents the disparity attention module proposed in Section \ref{sec:global_disp_att_module}. $\hat D^k$ and $D_{disp}^{k}$ denote the corresponding estimated depth and predicted disparity, respectively.

\paragraph{CameraNet}

{\small FBNetV3-B} \cite{dai2021fbnetv3}, without a classifier head, is utilized as the encoder. To feed video clips consisting of target and reference frames to the network, the number of channels in the first convolutional layer is changed from $3$ to $9$. The encoded high-level semantic features are first squeezed by a $1 \times 1$ filter with a stride of 1 and then transformed by employing  
two $3\times 3$ convolution layers followed by a rectified linear unit (ReLU) function. Finally, the relative pose, which is parameterized with 6 degrees of freedom (DOFs) whose first three dimensions denote translation and whose last three dimensions denote the rotation vector, is regressed by aggregating the estimation values at all spatial locations via global average pooling.

\vspace{-0.2cm}
\section{Experiments}\label{sec:experiments}
\subsection{Dataset}
We conducted experiments on the KITTI RAW dataset \cite{geiger2013vision}, the KITTI Odometry dataset \cite{geiger2012we} and the DDAD dataset \cite{guizilini20203d}. Similar to previous related works \cite{zhou2017unsupervised,yin2018geonet,ranjan2019competitive,godard2017unsupervised}, the KITTI RAW dataset was split as in \cite{eigen2014depth}, with approximately 40k frames used for training and 5k frames used for validation. We evaluated DepthNet on test data consisting of 697 test frames in accordance with Eigen's testing split, and we evaluated CameraNet on sequences $09-10$ of the KITTI Odometry dataset. The DDAD dataset was split as in \cite{guizilini20203d},
where 150 scene videos from the front camera were used for training and 50 scene videos were used for evaluation purposes.

\vspace{-0.5cm}
\subsection{Training Details}
The proposed learning framework was implemented using the PyTorch Library \cite{paszke2017automatic}. The training set was augmented using ColorJitter by randomly changing the brightness, contrast, saturation, and hue of each image, as well as performing random scaling, random horizontal flipping, and random cropping. During the training process, three consecutive video frames were used as a training sample and fed to the model. The same hyperparameter setting as that in \cite{wang2022cbwloss} was adopted for the loss function. Rather than storing all intermediate activations of the entire computed graph for backward computation, we instead recomputed them during the backward pass to save memory. DepthNet and CameraNet were coupled by the loss function from \cite{wang2022cbwloss} and jointly trained on an RTX 3090Ti GPU from scratch using the AdamW \cite{loshchilov2017decoupled} optimizer. The model for high-resolution input images was fine-tuned by utilizing the weight obtained from the adjacent low-resolution experiments. With increasing resolution, the batch size and learning rate were gradually reduced. The details are shown in Table \ref{tab:Batch_LR}.

\begin{table}[htbp]\small%[h]%[!htbp] 
	\vspace{-2pt}	
	\setlength\tabcolsep{2pt}
	\centering	
	\begin{tabular}{lcccccccc} 
		\toprule
	   	     
	     \multicolumn{1}{l}{Dataset}&\multicolumn{1}{c}{DepthNet}&\multicolumn{1}{c}{CameraNet} & \multicolumn{1}{l}{Batchsize} &\multicolumn{1}{c}{IterNum}& \multicolumn{1}{c}{LR} \\
			 
		\hline

		\multicolumn{1}{l}{KITTIRAW}& $128\times416$&$128\times416$&32& 100,000& $1e^{-4}$\\

		\multicolumn{1}{l}{KITTIRAW}&{$192\times640$}&$128\times416$&24& 50,000& $0.5e^{-4}$\\
		\multicolumn{1}{l}{KITTIRAW}&{$256\times 832$}&$128\times416$&16& 50,000& $0.4e^{-4}$\\
		\multicolumn{1}{l}{KITTIRAW}&{$320\times 1024$}&$128\times416$&10& 50,000& $0.3e^{-4}$\\
		
		\multicolumn{1}{l}{KITTIRAW}&{$384\times 1280$}&$128\times416$&8& 50,000& $0.2e^{-4}$\\

		\multicolumn{1}{l}{DDAD}&{$384\times 640$}&$384\times640$&12& 100,000& $0.5e^{-4}$\\
		
		\bottomrule 
	\end{tabular}
	\caption{Parameter settings.}\label{tab:Batch_LR}
	\vspace{-10pt}
\end{table}

\vspace{-0.3cm}
\subsection{Comparison with the State-of-the-Art Methods}

%%%%%%%%%%%    80m  
\begin{table*}[htbp]\small%[htbp]%[!htbp] 
	%\vspace{-0.1cm}
	\setlength\tabcolsep{2pt}
	\centering	
	
	\begin{tabular}{lcccccccccccccccc} %需要10列
		\toprule 
		\multicolumn{1}{l}{\multirow{2}*{\footnotesize Method}}&
		%\multicolumn{1}{c}{\multirow{2}*{\footnotesize Year}}&
		
		\multicolumn{1}{c}{\multirow{2}*{\footnotesize DE}}&
		\multicolumn{1}{c}{\multirow{2}*{\footnotesize PE}}&
		
		\multicolumn{1}{c}{\multirow{2}*{\footnotesize Data}}&
		%\multicolumn{1}{c}{\multirow{2}*{\footnotesize Cap (m)}}&
		\multicolumn{1}{c}{\multirow{2}*{\footnotesize RES}} &
		%\multicolumn{1}{c}{\multirow{2}*{\footnotesize Multi-Fr.}} &
		\multicolumn{1}{c}{\multirow{2}*{\footnotesize Sup + Multi-Fr?}} &
		\multicolumn{4}{c}{\footnotesize Error$\downarrow$}& 
		\multicolumn{3}{c}{\footnotesize Accuracy$\uparrow$}\\
		\multicolumn{6}{c}{}&\footnotesize AbsRel&\footnotesize SqRel&\footnotesize RMSE&\footnotesize RMSE log& \footnotesize $\delta_1$ &\footnotesize $ \delta_2$&\footnotesize $\delta_3$&	\\ 	 
		%\hline %绘制一条水平横线
		\midrule

		\multicolumn{1}{l}{Shu et al. \cite{shu2020feature}}&RN50&RN18&K&320$\times$1024 &M+FeatureNet&0.104&0.729&4.481&0.179& 0.893&0.965&\underline{0.984}&\\%2020ECCV&
		
		\multicolumn{1}{l}{Ruhkamp et al. \cite{ruhkamp2021attention}}&DRN-D-54&-&K+CS&192$\times$640&M+Multi-Fr&0.103&0.746&4.483&0.180& 0.894&0.965&{0.983}&\\%Dilated Residual Networks
		
		\multicolumn{1}{l}{Jung et al. \cite{jung2021fine}}&RN50&RN18&K&192$\times$640  &M+Semantic&0.102&{0.675}&4.393&0.178& 0.893&{0.966}&\underline{0.984}& \\%2021ICCV&

		\multicolumn{1}{l}{Guizilini et al. \cite{guizilini2020semantically}}&PackNet&PN7&K&384$\times$1280&M+Semantic&0.100&0.761&{4.270}&\underline{0.175}& {0.902}&0.965&0.982&\\%2020ICLR

		\multicolumn{1}{l}{Watson et al. \cite{watson2021temporal}}&RN18&RN18&K+CS&192$\times$640&M+Multi-Fr&\underline{0.098}&0.770&4.459&{0.176}& 0.900&0.965&{0.983}&\\%2021CVPR&

		\multicolumn{1}{l}{Guizilini et al. \cite{guizilini2022multi}}&Depthformer&RN18&K+CS&192$\times$640&M+Multi-Fr&\textbf{0.090}&\underline{0.661}&\underline{4.149}&\underline{0.175}&\underline{0.905}&\underline{0.967}&\underline{0.984}&\\%&2022CVPR
		
		\multicolumn{1}{l}{Zhou et al. \cite{zhou2022self}}&Swin&-&K&384$\times$1280&S&\textbf{0.090}&\textbf{0.538}&\textbf{3.896}&\textbf{0.169}&\textbf{0.906}&\textbf{0.969}&\textbf{0.985}&\\

		\midrule  %  以下是128x416

		\multicolumn{1}{l}{Zhang et al. \cite{zhang2020unsupervised}}&DenseDepthNet&PN7&K&128$\times$416  &M&0.137&0.972&5.367&0.219& 0.815&0.940&0.976& \\%2020TIP

		%\multicolumn{1}{l}{Zhang et al. \cite{zhang2020unsupervised}}&2020TIP&K+CS&128$\times$416  &M&0.133&0.968&5.227&0.216& &0.831&0.943&0.976& \\

		\multicolumn{1}{l}{Li et al. \cite{li2020unsupervised}}&DN&RN18&K&128$\times$416 &M&0.130&0.950&5.138&0.209 &0.843&0.948&0.978&\\		% 2021CoRL

		%\multicolumn{1}{l}{Jia et al. \cite{jia2021self} v2} &2022TITS&K&128$\times$416 &M&0.130&0.957&4.907&0.203& &0.851&0.954&0.980& \\
		
		\multicolumn{1}{l}{Godard et al. \cite{godard2019digging}}&RN18&RN18&K&128$\times$416  &M&0.128&1.087&5.171&0.204& 0.855&0.953&0.978& \\%2019ICCV

		\multicolumn{1}{l}{Yan et al. \cite{yan2021channel}}&RN50&RN50&K&128$\times$416 &M&{0.116}&{0.893}&{4.906}&{0.192} &{0.874}&{0.957}&{0.981}& \\%2021 3DV

		\multicolumn{1}{l}{\textbf{HQDec (Ours)}}&EffV2s&FBv3&K&128$\times$416  &M&\underline{0.103}&\underline{0.706}&\underline{4.569}&\underline{0.176}&\underline{0.882}&\underline{0.962}&\underline{0.985}& \\
		
		\multicolumn{1}{l}{\textbf{HQDec (Ours)$^\ddagger$}}&EffV2s&FBv3&K&128$\times$416  &M&\textbf{0.099}&\textbf{0.693}&\textbf{4.494}&\textbf{0.173}& \textbf{0.887}&\textbf{0.963}&\textbf{0.985}& \\

		\midrule % 以下是  192x640

		\multicolumn{1}{l}{Zhang et al.\cite{zhang2022self}}&\added{RN18}&\added{RN18}&K&192$\times$640  &M&0.112&0.856&4.778&0.190& 0.880&0.961&0.982& \\%2022TIP
		
		\multicolumn{1}{l}{Guizilini et al. \cite{guizilini20203d}}&\added{PackNet}&\added{PN7*}&K&192$\times$640  &M&0.111&0.785&4.601&0.189& 0.878&0.960&0.982& \\%2020CVPR

		\multicolumn{1}{l}{Song et al. \cite{song2021mlda}}&\added{MLDANet}&-&K&192$\times$640  &M&0.110&0.824&4.632&0.187 &0.883&0.961&0.982& \\%&2021TIP
		
		\multicolumn{1}{l}{Lyu et al. \cite{lyu2021hr}}&\added{RN18}&\added{RN18}&K&192$\times$640  &M&0.109&0.792&4.632&0.185& 0.884&0.962&0.983& \\%&2021AAAI

		\multicolumn{1}{l}{Johnston et al. \cite{johnston2020self}}&\added{RN101}&-&K&192$\times$640  &M&0.106&0.861&4.699&0.185& 0.889&0.962&0.982& \\%2020CVPR

		\multicolumn{1}{l}{Yan et al. \cite{yan2021channel}}&\added{RN50}&\added{RN50}&K&192$\times$640 &M&0.105&0.769&4.535&0.181& 0.892&0.964&0.983& \\%2021 3DV

		\multicolumn{1}{l}{Petrovai et al. \cite{petrovai2022exploiting}}&\added{RN50}&\added{RN18}&K+CS&192$\times$640  &M&{0.100}&{0.661}&{4.264}&{0.172} &{0.896}&\underline{0.967}&\textbf{0.985}& \\%&2022CVPR
		
		\multicolumn{1}{l}{\added{Zhou et al.} \cite{zhou2021self}}&\added{HR18}&\added{RN18}&\added{K}&\added{192$\times$640}&\added{M}&\added{0.102}&\added{0.764}&\added{4.483}&\added{0.180} &\added{0.896}&\added{0.965}&\added{0.983}& \\%%added{2021 BMVC}
		
		\multicolumn{1}{l}{\added{Zhao et al.} \cite{zhao2022monovit}}&\added{MPVit}&\added{RN18}&\added{K}&\added{192$\times$640}&\added{M}&\added{0.099}&\added{0.708}&\added{4.372}&\added{0.175}& \added{\underline{0.900}}&\added{\underline{0.967}}&\added{\underline{0.984}}& \\%\added{2022 3DV}
		
		\multicolumn{1}{l}{\added{He et al.} \cite{he2022ra}}&\added{HR18}&\added{RN18}&\added{K}&\added{192$\times$640} &\added{M}&\added{\underline{0.096}}&\added{\textbf{0.632}}&\added{\textbf{4.216}}&\added{0.171}& \added{\textbf{0.903}}&\added{\textbf{0.968}}&\added{\textbf{0.985}}& \\%&\added{2022 ECCV}
		
		\multicolumn{1}{l}{\added{Han et al.} \cite{han2022transdssl}}&\added{Swin}&\added{PN7}&\added{K}&\added{192$\times$640} &\added{M}&\added{0.098}&\added{0.728}&\added{4.458}&\added{0.176}& \added{0.898}&\added{0.966}&\added{\underline{0.984}}& \\%\added{2022 RAL}

		\multicolumn{1}{l}{\textbf{\added{HQDec (Ours)}}}&\added{EffV2s}&\added{FBv3}&\added{K}&\added{192$\times$640} &\added{M}&\added{\underline{0.096}}&\added{0.654}&\added{4.281}&\added{\underline{0.169}}& \added{0.896}&\added{0.965}&\added{\textbf{0.985}}& \\
		
		\multicolumn{1}{l}{\textbf{HQDec (Ours)\added{$^\ddagger$}}}&\added{EffV2s}&\added{FBv3}&K&192$\times$640  &M&\textbf{0.092}&\underline{0.642}&\underline{4.233}&\textbf{0.167}& {0.899}&\underline{0.966}&\textbf{0.985}& \\

		\midrule  % 以下是  256x832

		\multicolumn{1}{l}{Bian et al.\cite{bian2019unsupervised}}&\added{DRN}&\added{PN7}&K&256$\times$832 &M&0.137&1.089&5.439&0.217& 0.830&0.942&0.975&\\%2019NeurIPS
		
		\multicolumn{1}{l}{Jia et al.\cite{jia2021self}}&\added{RN18}&\added{PN7}&K&256$\times$832 &M&0.136&0.895&4.834&0.199& 0.832&0.950&\underline{0.982}&\\%2022TITS
		
		\multicolumn{1}{l}{Bian et al.\cite{bian2019unsupervised}}&\added{DRN}&\added{PN7}&K+CS&256$\times$832 &M&0.128&1.047&5.234&0.208& 0.846&0.947&0.976&\\%2019NeurIPS

		\multicolumn{1}{l}{Wang et al.\cite{wang2022cbwloss}}&\added{RN50}&\added{PN7}&K+CS&256$\times$832 &M&{0.110}&{0.847}&{4.654}&{0.189}& \underline{0.882}&\underline{0.960}&0.981&\\
		
		\multicolumn{1}{l}{\textbf{\added{HQDec (Ours)}}}&\added{EffV2s}&\added{FBv3}&\added{K}&\added{256$\times$832}  &\added{M}&\added{\underline{0.094}}&\added{\underline{0.655}}&\added{\underline{4.156}}&\added{\underline{0.167}}& \added{\underline{0.905}}&\added{\textbf{0.967}}&\added{\textbf{0.985}}& \\
		
		\multicolumn{1}{l}{\textbf{HQDec (Ours)\added{$^\ddagger$}}}&\added{EffV2s}&\added{FBv3}&K&256$\times$832  &M&\textbf{0.089}&\textbf{0.641}&\textbf{4.106}&\textbf{0.165}& \textbf{0.908}&\textbf{0.967}&\textbf{0.985}& \\

		\midrule  %以下是  320x1024
		\multicolumn{1}{l}{Godard et al.\cite{godard2019digging}}&\added{RN18}&\added{RN18}&K&320$\times$1024 &M&0.115&0.882&4.701&0.190& 0.879&0.961&0.982&\\%2019ICCV
		
		\multicolumn{1}{l}{Song et al. \cite{song2021mlda}}&\added{MLDANet}&-&K&320$\times$1024  &M&0.110&0.790&4.579&0.185& 0.882&0.963&0.983& \\ %2021TIP

		\multicolumn{1}{l}{Lyu et al. \cite{lyu2021hr}}&\added{RN18}&\added{RN18}&K&320$\times$1024  &M&0.106&0.755&4.472&0.181& 0.892&{0.966}&{0.984}& \\%2021AAAI

		\multicolumn{1}{l}{\added{Masoumian et al.} \cite{masoumian2023gcndepth}}&\added{RN50}&\added{RN18}&\added{K}&\added{320$\times$1024}  &\added{M}&\added{0.104}&\added{0.720}&\added{4.494}&\added{0.181}&\added{0.888}&\added{0.965}&\added{0.984}& \\%

		\multicolumn{1}{l}{Yan et al. \cite{yan2021channel}}&\added{RN50}&\added{RN50}&K&320$\times$1024  &M&0.102&0.734&4.407&0.178& 0.898&{0.966}&{0.984}& \\%2021 3DV

		\multicolumn{1}{l}{Petrovai et al. \cite{petrovai2022exploiting}}&\added{RN50}&\added{RN18}&K+CS&320$\times$1024&M&{0.098}&{0.674}&{4.187}&{0.170}& {0.902}&\textbf{\textbf{0.968}}&\underline{0.985}& \\%2022CVPR
		
		\multicolumn{1}{l}{\added{Zhou et al.} \cite{zhou2021self}}&\added{HR18}&\added{RN18}&\added{K}&\added{320$\times$1024}&\added{M}&\added{0.097}&\added{0.722}&\added{4.345}&\added{0.174}& \added{0.907}&\added{\underline{0.967}}&\added{0.984}& \\%\added{2021 BMVC}
		
		\multicolumn{1}{l}{\added{Zhao et al.} \cite{zhao2022monovit}}&\added{MPVit}&\added{RN18}&\added{K}&\added{320$\times$1024}&\added{M}&\added{0.096}&\added{0.714}&\added{4.292}&\added{0.172}& \added{\underline{0.908}}&\added{\textbf{0.968}}&\added{0.984}& \\%\added{2022 3DV}

		\multicolumn{1}{l}{\textbf{\added{HQDec (Ours)}}}&\added{EffV2s}&\added{FBv3}&\added{K}&\added{320$\times$1024}  &\added{M}&\added{\underline{0.093}}&\added{\underline{0.654}}&\added{\underline{4.102}}&\added{\underline{0.165}}& \added{0.906}&\added{\textbf{0.968}}&\added{\textbf{0.986}}& \\
		
		\multicolumn{1}{l}{\textbf{HQDec (Ours)\added{$^\ddagger$}}}&\added{EffV2s}&\added{FBv3}&K&320$\times$1024&M&\textbf{0.088}&\textbf{0.638}&\textbf{4.052}&\textbf{0.163}& \textbf{0.909}&\textbf{0.968}&\underline{0.985}& \\

		\midrule % 384x1280
		
		\multicolumn{1}{l}{Guizilini et al. \cite{guizilini20203d}}&\added{PackNet}&\added{PN7*}&K&384$\times$1280 &M&0.107&0.802&4.538&0.186& 0.889&0.962&0.981&\\%2020CVPR

		\multicolumn{1}{l}{Lyu et al. \cite{lyu2021hr}}&\added{RN18}&\added{RN18}&K&384$\times$1280  &M&0.104&0.727&4.410&0.179& 0.894&\underline{0.966}&\underline{0.984}& \\%2021AAAI
		
		\multicolumn{1}{l}{Wang et al. \cite{wang2022cbwloss}}&\added{RN50}&\added{PN7}&K+CS&384$\times$1280  &M&0.104&0.798&4.501&0.184& 0.889&0.961&0.982& \\
		
		\multicolumn{1}{l}{Yan et al. \cite{yan2021channel}}&\added{RN50}&\added{RN50}&K&384$\times$1280  &M&{0.102}&{0.715}&{4.312}&{0.176}& {0.900}&\underline{0.968}&\underline{0.984}& \\%2021 3DV
		
		 \multicolumn{1}{l}{\added{Zhao et al.} \cite{zhao2022monovit}}&\added{MPVit}&\added{RN18}&\added{K}&\added{384$\times$1280}&\added{M}&\added{0.094}&\added{0.682}&\added{4.200}&\added{0.170}& \added{\textbf{0.912}}&\added{\textbf{0.969}}&\added{0.984}& \\%\added{2022 3DV}

		\multicolumn{1}{l}{\textbf{\added{HQDec (Ours)}}}&\added{EffV2s}&\added{FBv3}&\added{K}&\added{384$\times$1280}  &\added{M}&\underline{\added{0.092}}&\underline{\added{0.634}}&\underline{\added{4.079}}&\textbf{\added{0.164}}& {\added{0.908}}&\textbf{\added{\underline{0.968}}}&\textbf{\added{0.986}}& \\
		
		\multicolumn{1}{l}{\textbf{HQDec (Ours)\added{$^\ddagger$}}}&\added{EffV2s}&\added{FBv3}&K&384$\times$1280  &M&\textbf{0.088}&\textbf{0.624}&\textbf{4.031}&\textbf{0.162}& \underline{0.911}&\underline{0.968}&\textbf{0.986}& \\

		\bottomrule %添加表格底部粗线
	\end{tabular}
	\caption{Performance comparison conducted on the KITTI dataset with monocular depth estimation capped at 80 m. \added{The prediction results were aligned by the median ground-truth LiDAR information.} `K' denotes that the models were trained only on KITTI, and CS\deleted{/IN+K} means that the models were fine-tuned on KITTI after pretraining them on the Cityscapes\deleted{/ImageNet} dataset. `M/\added{S}' refers to methods that were trained by using monocular (M)\added{/stereo   (S)} image pairs and inferred the depth from a single image. `Multi-Fr.' indicates that the depth map was predicted by utilizing multiple frames during the inference process. \added{`$^\ddagger$' indicates that the prediction results were aligned by the proposed adaptive scale alignment technique. `DE/PE' refers to the backbone of the encoder used in the depth/pose estimation network. `RN18/RN50/RN101/DRN-D-54/Effv2s/HR18/MPVit/Swin-s/FBv3/' refer to the corresponding encoders based on ResNet18\cite{he2016deep}/ResNet50\cite{he2016deep}/ResNet101\cite{he2016deep}/dilated residual networks\cite{yu2017dilated}/EfficientNetv2s\cite{tan2021efficientnetv2}/HR18\cite{wang2020deephr}/MPVit\cite{lee2022mpvit}/the Swin transformer\cite{liu2021swin}/FBNetV3-B\cite{dai2021fbnetv3}, respectively. `PN7' refers to a simple pose estimation network consisting of seven convolution layers that was designed in \cite{zhou2017unsupervised}. `*' indicates that group normalization was used after each convolution layer in PN7.  `PackNet/Depthformer/DN/DRN/DenseDepthNet/MLDANet/' denote the corresponding networks in \cite{guizilini20203d}/\cite{guizilini2022multi}/\cite{zhou2017unsupervised}/\cite{ranjan2019competitive}/\cite{zhang2020unsupervised}/\cite{song2021mlda}, respectively.}} \label{tab:depth_compared_previous_method_80m}
	\vspace{-10pt}
\end{table*}

\begin{table*}[htbp]\small%[htbp]%[!htbp] 
	%\vspace{-0.1cm}
	\setlength\tabcolsep{2pt}
	\centering

	\begin{tabular}{lcccccccccccccccc} %需要10列
		\toprule %添加表格头部粗线
		\multicolumn{1}{l}{\multirow{2}*{\footnotesize Method}}&
		%\multicolumn{1}{c}{\multirow{2}*{\footnotesize Year}}&
		\multicolumn{1}{c}{\multirow{2}*{\footnotesize DE}}&
		\multicolumn{1}{c}{\multirow{2}*{\footnotesize PE}}&
		
		\multicolumn{1}{c}{\multirow{2}*{\footnotesize Data}}&
		%\multicolumn{1}{c}{\multirow{2}*{\footnotesize Cap (m)}}&
		\multicolumn{1}{c}{\multirow{2}*{\footnotesize RES}} &
		%\multicolumn{1}{c}{\multirow{2}*{\footnotesize Multi-Fr.}} &
		\multicolumn{1}{c}{\multirow{2}*{\footnotesize Sup + Multi-Fr?}} &
		\multicolumn{4}{c}{\footnotesize Error$\downarrow$}&
		\multicolumn{3}{c}{\footnotesize Accuracy$\uparrow$}\\
		\multicolumn{6}{c}{}&\footnotesize AbsRel&\footnotesize SqRel&\footnotesize RMSE&\footnotesize RMSE log& \footnotesize $\delta_1$ &\footnotesize $ \delta_2$&\footnotesize $\delta_3$&	\\

		\midrule
		\multicolumn{1}{l}{Watson et al. \cite{watson2021temporal}}&\added{RN18}&\added{RN18}&K+CS&192$\times$640  &M+Multi-Fr&\underline{0.064}&0.320&3.187&0.104& 0.946&0.990&\underline{0.995}& \\%2021CVPR

		\multicolumn{1}{l}{Guizilini et al. \cite{guizilini2022multi}}&\added{Depthformer}&\added{RN18}&K+CS&192$\times$640  &M+Multi-Fr&\textbf{0.055}&\underline{0.271}&\underline{2.917}&\underline{0.095}& \underline{0.955}&\underline{0.991}&\textbf{0.998}& \\%2022CVPR
		
		\multicolumn{1}{l}{Guizilini et al. \cite{guizilini2022multi}}&\added{Depthformer}&\added{RN18}&K+CS&352$\times$1216  &M+Multi-Fr&\textbf{0.055}&\textbf{0.265}&\textbf{2.723}&\textbf{0.092}& \textbf{0.959}&\textbf{0.992}&\textbf{0.998}& \\

		\midrule		
		
		\multicolumn{1}{l}{Bian et al. \cite{bian2019unsupervised}}&\added{DRN}&\added{PN7}&K+CS&256$\times$832  &M&0.098&0.650&4.398&0.153& 0.892&0.972&0.991& \\%2019NeurIPS

		\multicolumn{1}{l}{Godard et al. \cite{godard2019digging}}&\added{RN18}&\added{RN18}&K&192$\times$640  &M&0.090&0.545&3.942&0.137& 0.914&0.983&0.995& \\%2019ICCV
		
		\multicolumn{1}{l}{Guizilini et al. \cite{guizilini20203d}}&\added{PackNet}&\added{PN7*}&K+CS&192$\times$640  &M&0.078&0.420&3.485&0.121& 0.931&0.986&0.996& \\%2020CVPR
		
		\multicolumn{1}{l}{\added{Zhou et al. }\cite{zhou2021self}}&\added{HR18}&\added{RN18}&\added{K}&\added{192$\times$640 } &\added{M}&\added{0.076}&\added{0.414}&\added{3.493}&\added{0.119}& \added{0.936}&\added{0.988}&\added{\underline{0.997}}& \\%\added{2021BMVC}
		
		\multicolumn{1}{l}{\added{Zhao et al. }\cite{zhao2022monovit}}&\added{MPVit}&\added{RN18}&\added{K}&\added{192$\times$640 } &\added{M}&\added{0.075}&\added{0.389}&\added{3.419}&\added{0.115}& \added{0.938}&\added{0.989}&\added{\underline{0.997}}& \\%\added{2022}
		
		\multicolumn{1}{l}{\added{He et al. }\cite{he2022ra}}&\added{HR18}&\added{RN18}&\added{K}&\added{192$\times$640 } &\added{M}&\added{0.074}&\added{0.362}&\added{3.345}&\added{0.114}& \added{0.940}&\added{0.990}&\added{\underline{0.997}}& \\%\added{2022ECCV}

		\multicolumn{1}{l}{Wang et al. \cite{wang2022cbwloss}}&\added{RN50}&\added{PN7}&K+CS&256$\times$832  &M&0.077&0.425&3.537&0.121& 0.934&0.985&0.996& \\

		\multicolumn{1}{l}{\textbf{\added{HQDec (Ours)}}}&\added{EffV2s}&\added{FBv3}&\added{K}&\added{128$\times$416}  &\added{M}&\added{0.074}&\added{0.403}&\added{3.746}&\added{0.121}& \added{0.930}&\added{0.986}&\added{\underline{0.997}}& \\
		
		\multicolumn{1}{l}{\textbf{HQDec (Ours)\added{$^\ddagger$}}}&\added{EffV2s}&\added{FBv3}&K&128$\times$416  &M&0.071&0.384&3.632&0.117& 0.935&0.987&\underline{0.997}& \\

		\multicolumn{1}{l}{\textbf{\added{HQDec (Ours)}}}&\added{EffV2s}&\added{FBv3}&\added{K}&\added{192$\times$640}  &\added{M}&\added{0.065}&\added{0.328}&\added{3.289}&\added{0.107}& \added{0.945}&\added{0.990}&\added{\underline{0.997}}& \\
		
		\multicolumn{1}{l}{\textbf{HQDec (Ours)\added{$^\ddagger$}}}&\added{EffV2s}&\added{FBv3}&K&192$\times$640  &M&{0.062}&{0.318}&{3.231}&{0.105}& {0.948}&{0.990}&\underline{0.997}& \\

		\multicolumn{1}{l}{\textbf{\added{HQDec (Ours)}}}&\added{EffV2s}&\added{FBv3}&\added{K}&\added{256$\times$832}  &\added{M}&\added{\underline{0.060}}&\added{\underline{0.298}}&\added{\underline{3.004}}&\added{\underline{0.100}}& \added{\underline{0.955}}&\added{\underline{0.991}}&\added{\textbf{0.998}}& \\
		\multicolumn{1}{l}{\textbf{HQDec (Ours)\added{$^\ddagger$}}}&\added{EffV2s}&\added{FBv3}&K&256$\times$832  &M&\textbf{0.058}&\textbf{0.289}&\textbf{2.953}&\textbf{0.098}& \textbf{0.958}&\textbf{0.992}&\textbf{0.998}& \\

		\midrule
		\multicolumn{1}{l}{Godard et al. \cite{godard2019digging}}&\added{RN18}&\added{RN18}&K&320$\times$1024  &M&0.086&0.462&3.577&0.127& 0.924&0.986&0.996& \\%2019ICCV

		\multicolumn{1}{l}{Lyu et al. \cite{lyu2021hr}}&\added{RN18}&\added{RN18}&K&384$\times$1280  &M&0.075&0.357&3.239&0.113& 0.937&\underline{0.991}&\textbf{0.998}& \\%2021AAAI

		\multicolumn{1}{l}{Guizilini et al. \cite{guizilini20203d}}&\added{PackNet}&\added{PN7*}&K+CS&384$\times$1280  &M&{0.071}&0.359&3.153&{0.109}& {0.944}&0.990&\underline{0.997}& \\%2020CVPR

		\multicolumn{1}{l}{Wang et al. \cite{wang2022cbwloss}}&\added{RN50}&\added{PN7}&K+CS&384$\times$1280  &M&0.072&0.377&3.358&0.115& 0.939&0.987&0.996& \\

		\multicolumn{1}{l}{\added{Zhao et al. }\cite{zhao2022monovit}}&\added{MPVit}&\added{RN18}&\added{K}&\added{384$\times$1280 } &\added{M}&\added{0.067}&\added{0.328}&\added{3.108}&\added{0.104}& \added{0.950}&\added{\textbf{0.992}}&\added{\textbf{0.998}}& \\%\added{2022}

		\multicolumn{1}{l}{\textbf{\added{HQDec (Ours)}}}&\added{EffV2s}&\added{FBv3}&\added{K}&\added{320$\times$1024}  &\added{M}&\added{0.061}&\added{0.296}&\added{2.944}&\added{\underline{0.099}}& \added{\underline{0.956}}&\added{\textbf{0.992}}&\added{\textbf{0.998}}& \\
		
		\multicolumn{1}{l}{\textbf{HQDec (Ours)\added{$^\ddagger$}}}&\added{EffV2s}&\added{FBv3}&K&320$\times$1024  &M&\textbf{0.058}&\underline{0.286}&\textbf{2.896}&\textbf{0.097}& \textbf{0.959}&\textbf{0.992}&\textbf{0.998}& \\

		\multicolumn{1}{l}{\textbf{\added{HQDec (Ours)}}}&\added{EffV2s}&\added{FBv3}&\added{K}&\added{384$\times$1280}  &\added{M}&\added{\underline{0.061}}&\added{0.292}&\added{2.976}&\added{\underline{0.099}}& \added{\underline{0.956}}&\added{\textbf{0.992}}&\added{\textbf{0.998}}& \\
		
		\multicolumn{1}{l}{\textbf{HQDec (Ours)\added{$^\ddagger$}}}&\added{EffV2s}&\added{FBv3}&K&384$\times$1280  &M&\textbf{0.058}&\textbf{0.284}&\underline{2.921}&\textbf{0.097}& \textbf{0.959}&\textbf{0.992}&\textbf{0.998}& \\

		\bottomrule %添加表格底部粗线
	\end{tabular}
	\caption{Monocular depth estimation performance comparison conducted on the improved  KITTI dataset \cite{uhrig2017sparsity}. \added{The prediction results were aligned by the median ground-truth LiDAR information.} \added{`$^\ddagger$' indicates that the prediction results were aligned by the proposed adaptive scale alignment technique.}}\label{tab:depth_compared_previous_method_80m_on_improved_kitti_data}
	\vspace{-6pt}
\end{table*}

%  ddad results
\begin{table*}[htbp]%\small%[htbp]%[!htbp] 	
	%\vspace{-0.1cm}
	\setlength\tabcolsep{2pt}
	\centering		
	\begin{tabular}{lccccccccccccccccccc} %需要10列
		\toprule %添加表格头部粗线
		\multicolumn{1}{l}{\multirow{2}*{\footnotesize Method}}&
		%\multicolumn{1}{c}{\multirow{2}*{\footnotesize Year}}&
		
		\multicolumn{1}{c}{\multirow{2}*{\footnotesize DE}}&
		\multicolumn{1}{c}{\multirow{2}*{\footnotesize PE}}&
		
		\multicolumn{1}{c}{\multirow{2}*{\footnotesize Sup + Multi-Fr?}} &		
		\multicolumn{1}{c}{\multirow{2}*{\footnotesize RES}}&
		\multicolumn{4}{c}{ Error$\downarrow$}&
		\multicolumn{3}{c}{\footnotesize Accuracy$\uparrow$}\\
		\multicolumn{5}{c}{}&\footnotesize AbsRel&\footnotesize SqRel&\footnotesize RMSE&\footnotesize RMSElog& \footnotesize $\delta_1$ &\footnotesize $ \delta_2$&\footnotesize $\delta_3$&	\\ 		
		\midrule	
		
		\multicolumn{1}{l}{{Guizilini et al.} \cite{guizilini20203d}}&\added{PackNet}&\added{PN7*}&M&{384$\times$640} &{0.162}&{3.917}&{13.452}&{0.269}& {0.823}&{-}&{-}&\\%2020CVPR

		\multicolumn{1}{l}{\added{Han et al.} \cite{han2022transdssl}}&\added{Swin}&\added{PN7}&\added{M}&\added{384$\times$640} &\added{0.151}&\added{3.591}&\added{14.350}&\added{0.244}&\added{-}&\added{-}&\added{-}&\\%2020CVPR
		
		\multicolumn{1}{l}{Guizilini et al. \cite{guizilini2021geometric}}&\added{RN101}&\added{RN18}&M+Semantic&384$\times$640&0.147&2.922&14.452&-& 0.809&-&-&\\%2021ICCV
		
		\multicolumn{1}{l}{Watson et al. \cite{watson2021temporal}}&\added{RN18}&\added{RN18}&M+Multi-Fr.&-&0.146&3.258&14.098&- &0.822&-&-&\\%2021CVPR
		
		\multicolumn{1}{l}{Guizilini et al. \cite{guizilini2022multi}}&\added{Depthformer}&\added{RN18}&M+Multi-Fr.&-&0.135&2.953&12.477&-& 0.836&-&-&\\%2022CVPR

		\multicolumn{1}{l}{Wang et al. \cite{wang2022cbwloss}}&\added{RN50}&\added{PN7}&M&384$\times$640&
	{0.121}&{1.229}&{6.721}&{0.187}& {0.853}&{0.958}&\textbf{0.985}&\\
		
		\multicolumn{1}{l}{\textbf{\added{HQDec (Ours)}}}&\added{EffV2s}&\added{FBv3}&\added{M}&\added{384$\times$640}&\added{\underline{0.113}}&\added{\underline{1.214}}&\added{\underline{6.286}}&\added{\underline{0.176}}& \added{\underline{0.876} }&\added{\underline{0.963}}&\added{\textbf{0.985}}&\\
		
		\multicolumn{1}{l}{\textbf{HQDec (Ours)\added{$^\ddagger$}}}&\added{EffV2s}&\added{FBv3}&M&384$\times$640&\textbf{0.107}&\textbf{1.173}&\textbf{6.109}&\textbf{0.171}& \textbf{0.885}&\textbf{0.964}&\textbf{0.985}&\\

		\bottomrule %添加表格底部粗线
	\end{tabular}	
	\caption{Monocular depth estimation performance comparison conducted on the DDAD dataset \cite{guizilini20203d}. \added{The prediction were aligned by the median ground-truth LiDAR information.} \added{`$^\ddagger$' indicates that the prediction results were aligned by the proposed adaptive scale alignment technique.}}\label{tab:depth_compared_on_ddad}
	\vspace{-0.6cm}
\end{table*}

In Table \ref{tab:depth_compared_previous_method_80m}, we quantitatively compare the proposed HQDec with the previously developed state-of-the-art methods at different resolutions. The results show that the proposed HQDec outperformed these state-of-the-art completely unsupervised monocular estimation methods \cite{bian2019unsupervised,jia2021self,zhang2020unsupervised,li2020unsupervised,godard2019digging,yan2021channel,zhang2022self,guizilini20203d,song2021mlda,lyu2021hr,johnston2020self,petrovai2022exploiting,wang2022cbwloss} and monocular estimation methods\label{key} \cite{klingner2020self,jung2021fine,shu2020feature,guizilini2020semantically} guided by semantic labels at the same resolution. The proposed HQDec, which only utilized a single frame to estimate the corresponding depth during inference, could achieve better performance that was similar to or even better than that of the multiframe approaches \cite{guizilini2022multi,watson2021temporal}, where multiframe images were required to predict the depth map during inference. In Table \ref{tab:depth_compared_previous_method_80m_on_improved_kitti_data}, we also provide the results evaluated on the improved KITTI ground-truth data \cite{uhrig2017sparsity}. These results also confirmed that the proposed HQDec outperformed all previously published self-supervised methods that estimate the corresponding depth map by utilizing only a single frame during inference. Benefitting from the estimated high-quality depth, we also achieved competitive pose estimation errors, although CameraNet and DepthNet were only trained on the KITTI RAW dataset, as shown in Table \ref{tab:camera_pose_previous_method}. To further verify the HQDec's generalization ability, we also evaluated the HQDec on the more challenging DDAD dataset \cite{guizilini20203d}. The results in Table \ref{tab:depth_compared_on_ddad} show that the proposed HQDec outperformed all previously published self-supervised methods, including the approaches that utilize multiple frames to predict the corresponding depth map and methods that use semantic labels as supervision signals. 

\begin{figure*}[t]
	\centering
	\scriptsize 	
	\rotatebox{90}{  {\footnotesize\textbf{ Ours}}\;
		{\footnotesize Wang}\quad
		{\footnotesize Lyu}\;
		{\footnotesize Guizilini}
		{\footnotesize Godard} 		
		{\footnotesize Input }
	}
\rotatebox{90}{\qquad\qquad\cite{wang2022cbwloss}
	\qquad\cite{lyu2021hr}
	\quad\cite{guizilini20203d}
	\qquad\cite{godard2019digging}\qquad GT}
\rotatebox{90}{ 38.38M
	 32.52M
	 14.62M
	 128.29M
	14.84M} 		
	\includegraphics[scale=0.065]{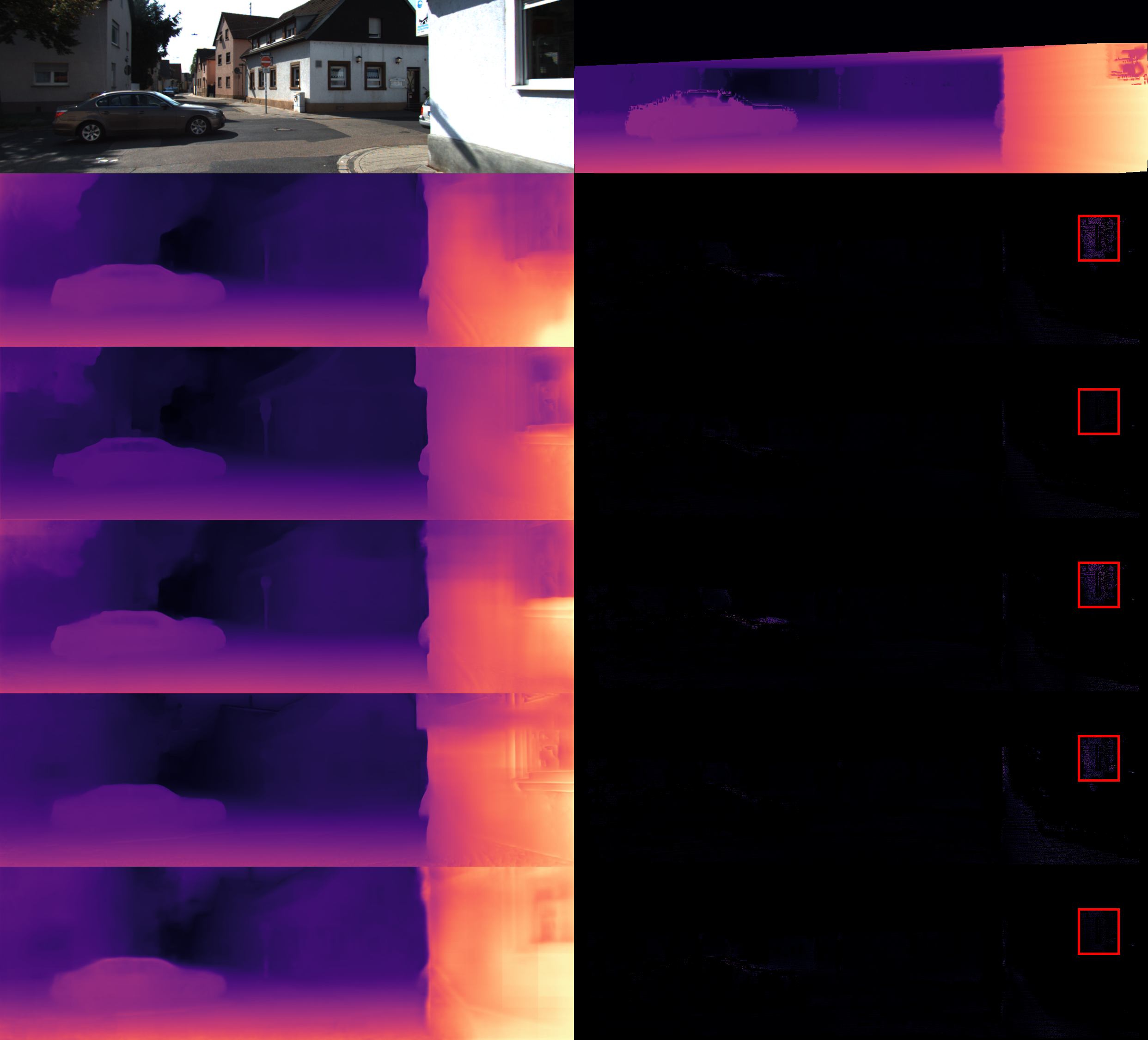}
	\includegraphics[scale=0.065]{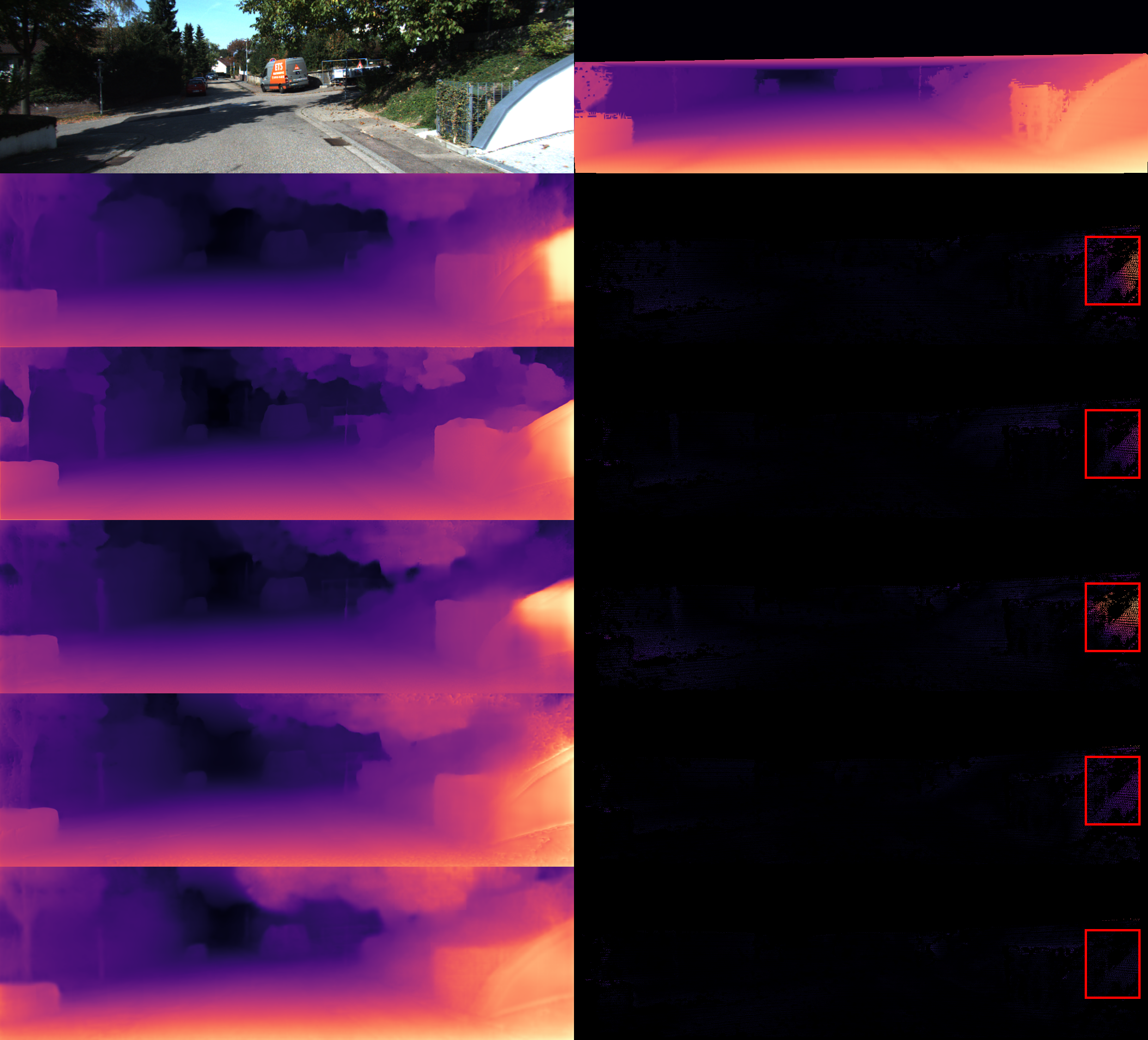}
	\includegraphics[scale=0.065]{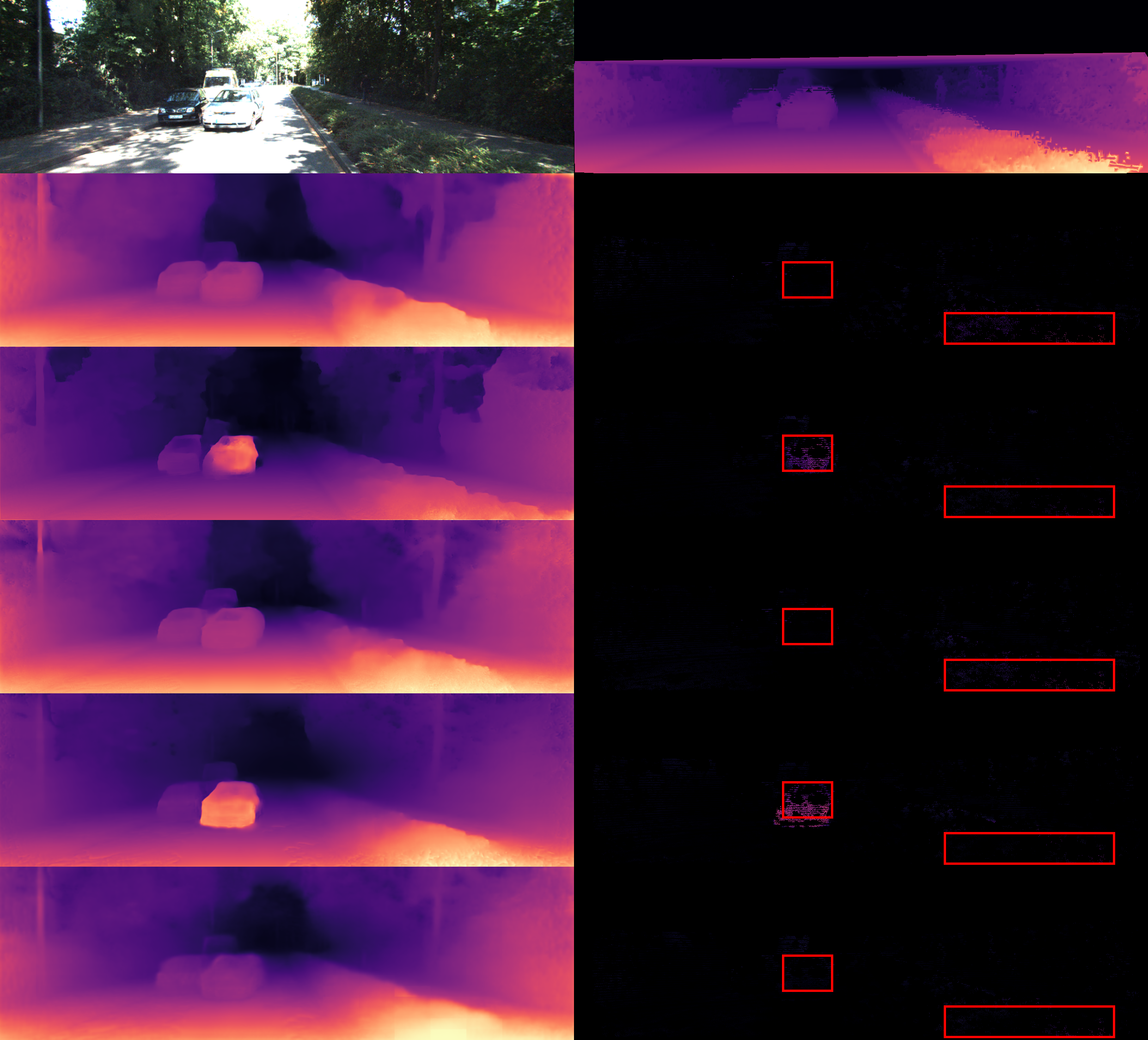}

	\caption{Qualitative comparison among the example results (384$\times$1280 or similar) obtained on the KITTI dataset. \added{The points without valid LIDAR measurements in the error maps were masked out by using the improved ground-truth maps \cite{uhrig2017sparsity}.}}\label{fig:compare_disp_kitti}%
	\vspace{-10pt}	
\end{figure*}

In Fig. \ref{fig:compare_disp_kitti}, we conduct a qualitative comparison with the previously developed methods \cite{godard2019digging,guizilini20203d,lyu2021hr,wang2022cbwloss} on some challenging samples selected from the KITTI dataset. In the left/middle samples, the previously developed methods \cite{godard2019digging,guizilini20203d,wang2022cbwloss,lyu2021hr} tended to inaccurately estimate the depths in some areas (e.g., the white wall in the left image and the white tent in the middle image). In the right sample, although the methods from \cite{godard2019digging,lyu2021hr} accurately estimated the depth of the white car, they failed to accurately predict the depth of the roadside lawn. The methods in \cite{wang2022cbwloss,guizilini20203d} exhibited the opposite phenomenon. However, the proposed HQDec could accurately predict the corresponding depths in these scenarios. We attribute this phenomenon to the fact that the proposed HQDec could utilize multilevel features with global and local context information to infer the current depth. The model was told that the same object (e.g., the white wall, the white tent, etc.) should have a similar depth by establishing long-range dependencies between the pixels. In addition, compared with the existing methods, the proposed HQDec could accurately predict not only the depth of the moving car by modeling the relationships between the moving car and its surrounding pixels but also the depth of the lawn in the right sample by modeling the long-range dependencies between the pixels in the lawn area.

\begin{figure}[htbp]
	%\vspace{-0.6cm}
	\includegraphics[scale=0.047]{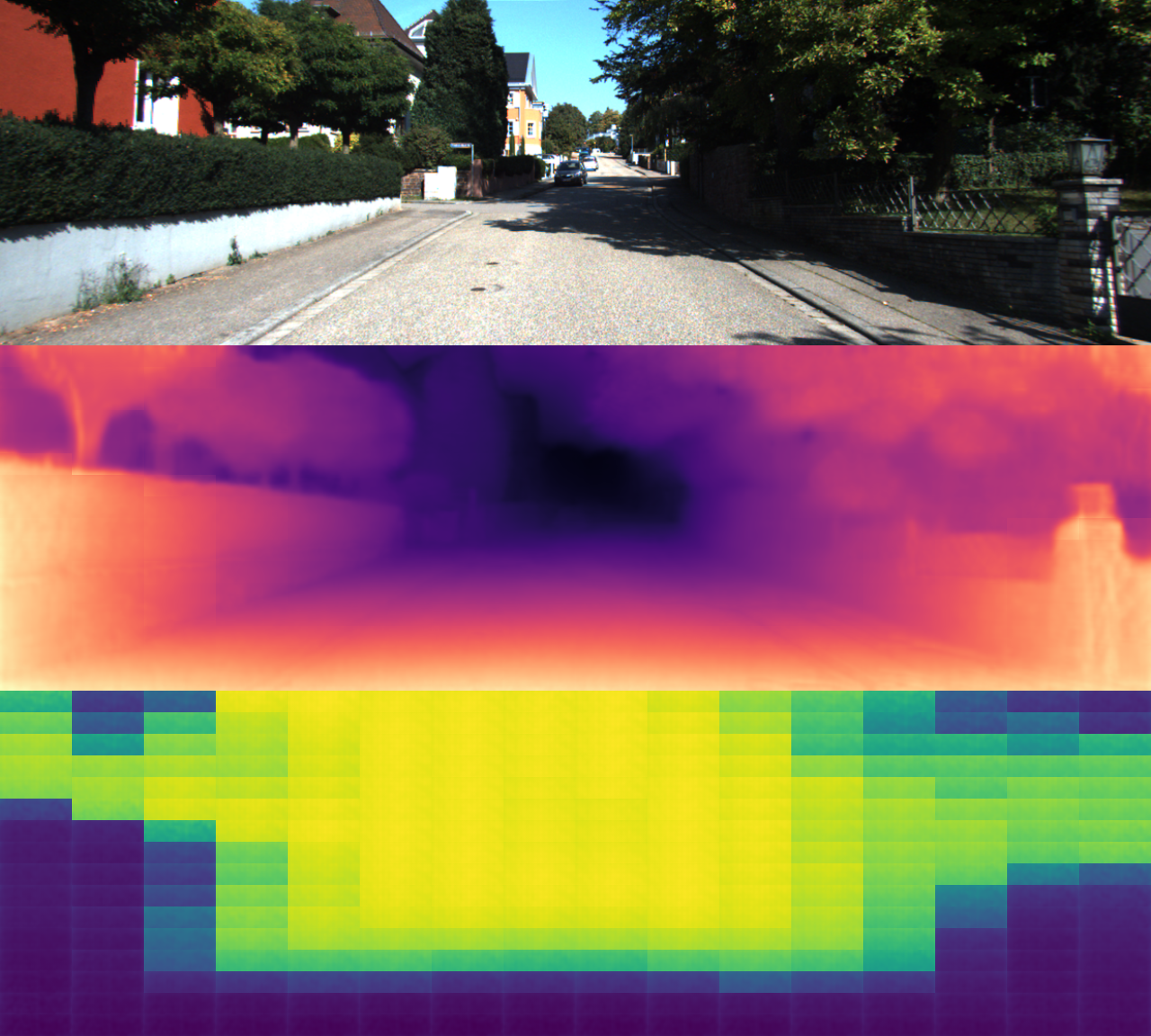}
	\includegraphics[scale=0.047]{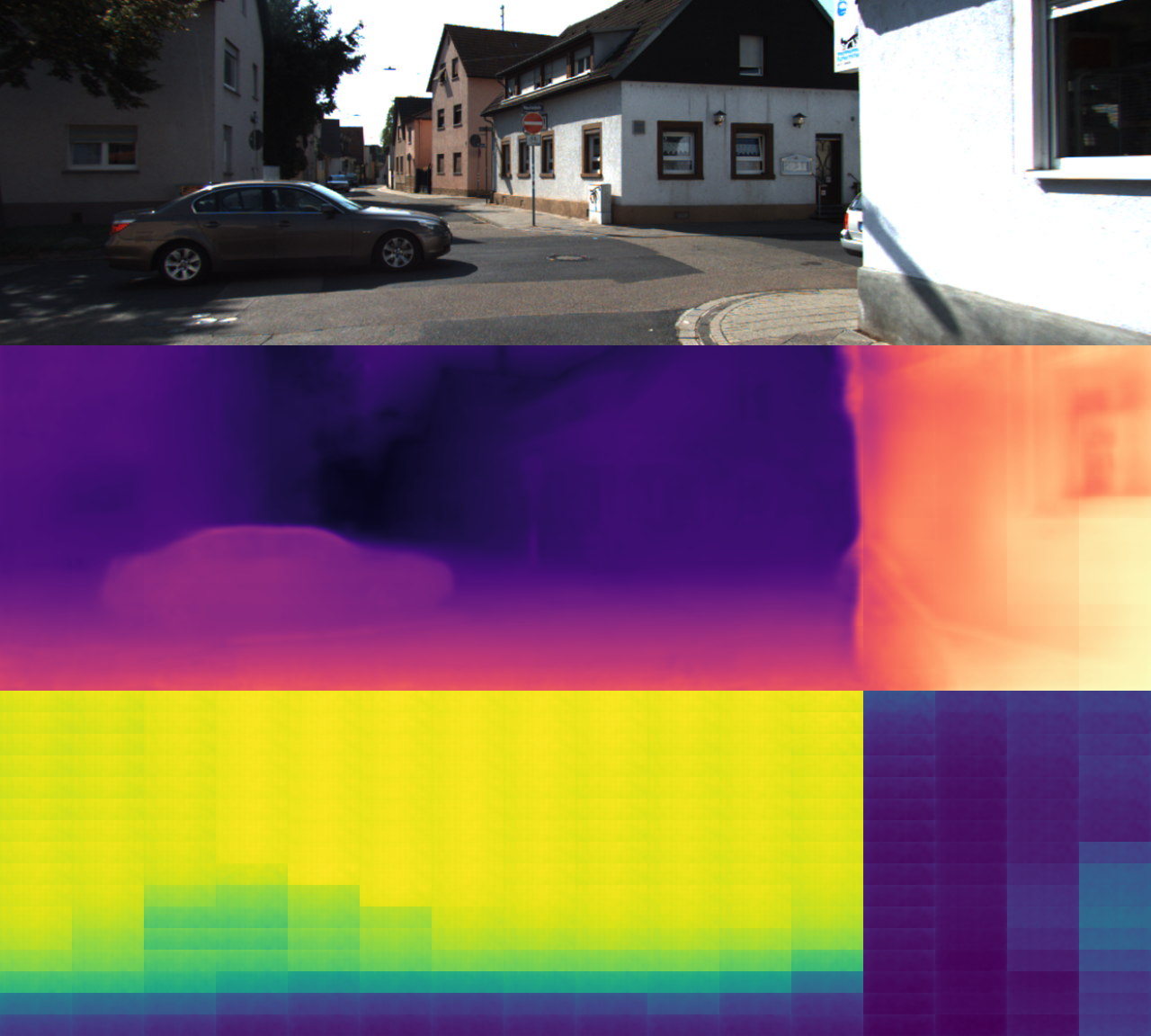}
	\includegraphics[scale=0.047]{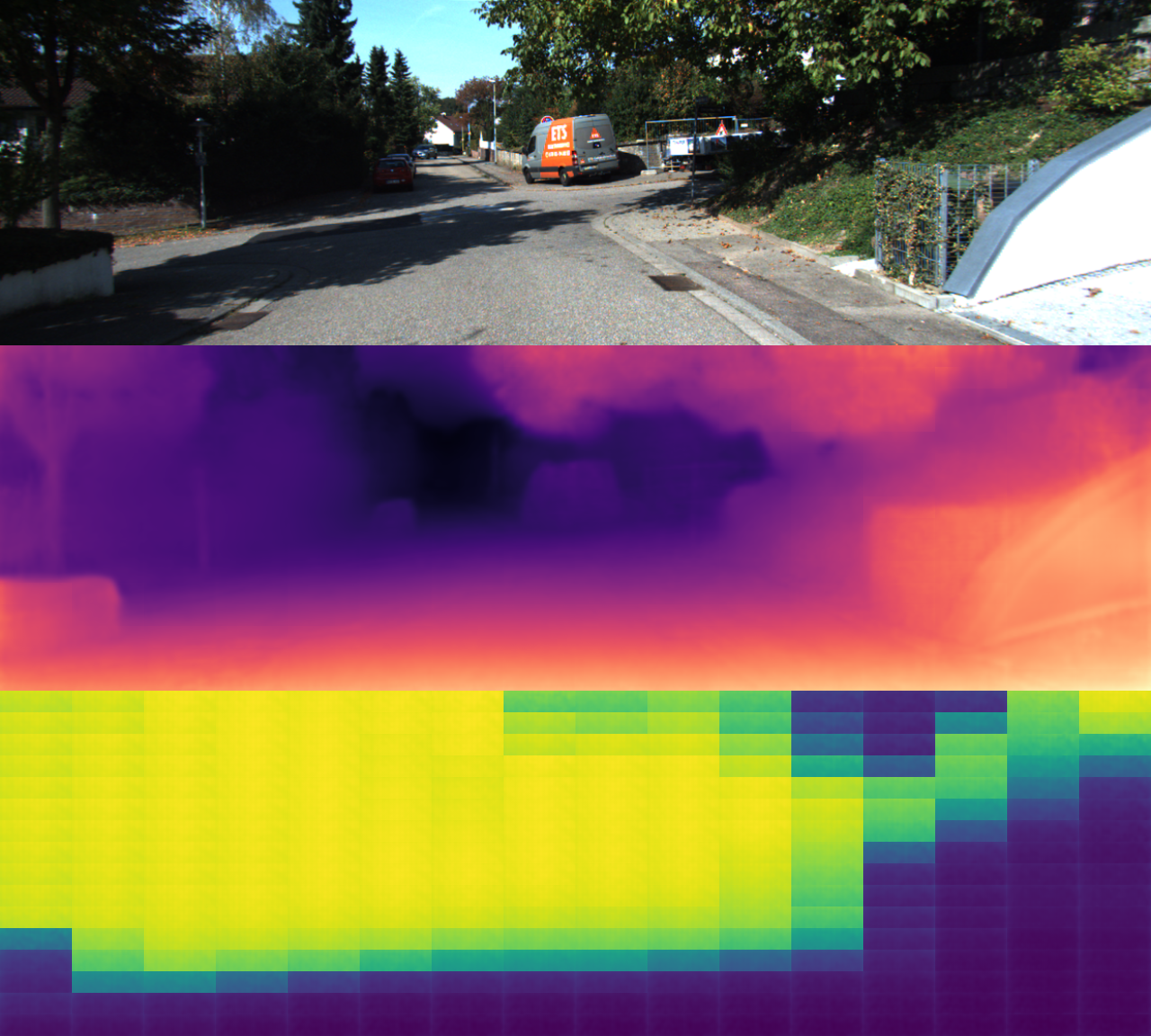}
	\includegraphics[scale=0.047]{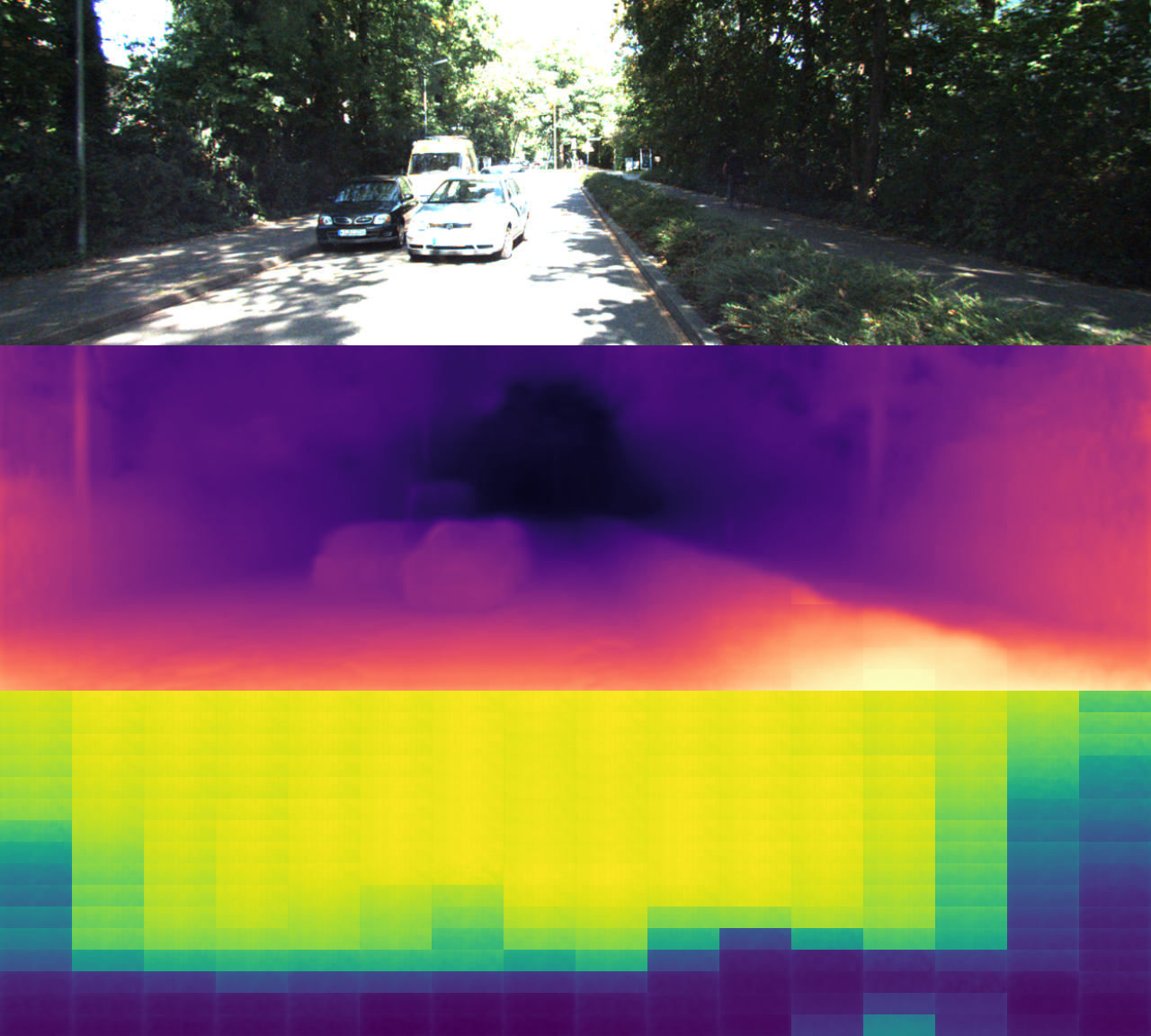}
	\caption{Visualization of the attention weights produced by the disparity attention module. }\label{fig:att_viz}
	\vspace{-0.8cm}
\end{figure}

\vspace{-0.01cm}
Fig. \ref{fig:att_viz} shows that areas in the scene that should theoretically have similar depths were given similar weights by the disparity attention module. 

\begin{figure*}[t]%htbp
	\centering
	\scriptsize 
	\rotatebox{90}{ $\ $ \textbf{Ours} $\ $  \quad
		Wang\cite{wang2022cbwloss} $\ $ 
		Guizilini\cite{guizilini20203d}$\ $	 	
		Input  
	}\rotatebox{90}{
		384$\tiny\times$640\quad\;
		384$\tiny\times$640\qquad
		384$\tiny\times$640\qquad
		GT
		
	}
	\includegraphics[scale=0.03]{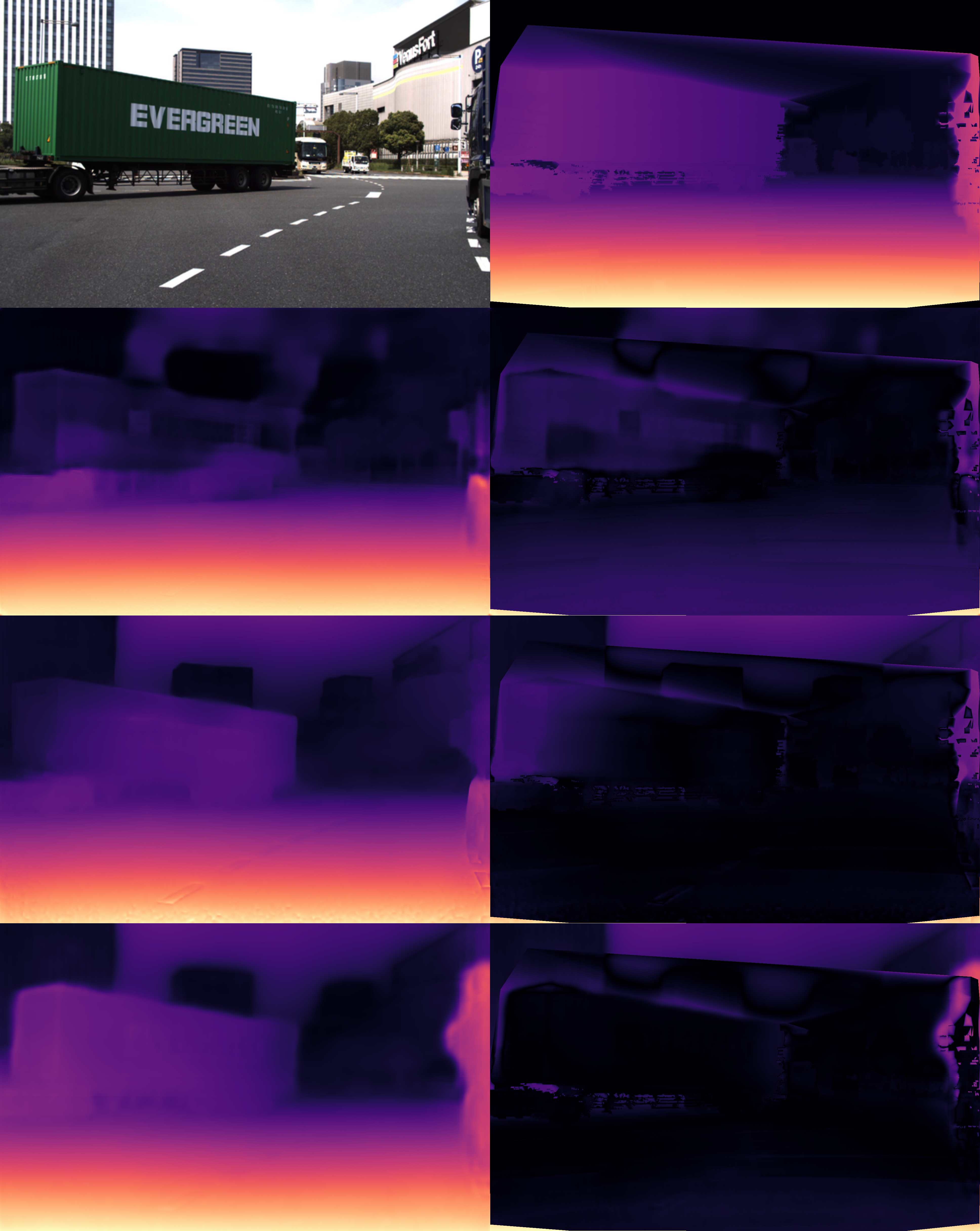}
	\includegraphics[scale=0.03]{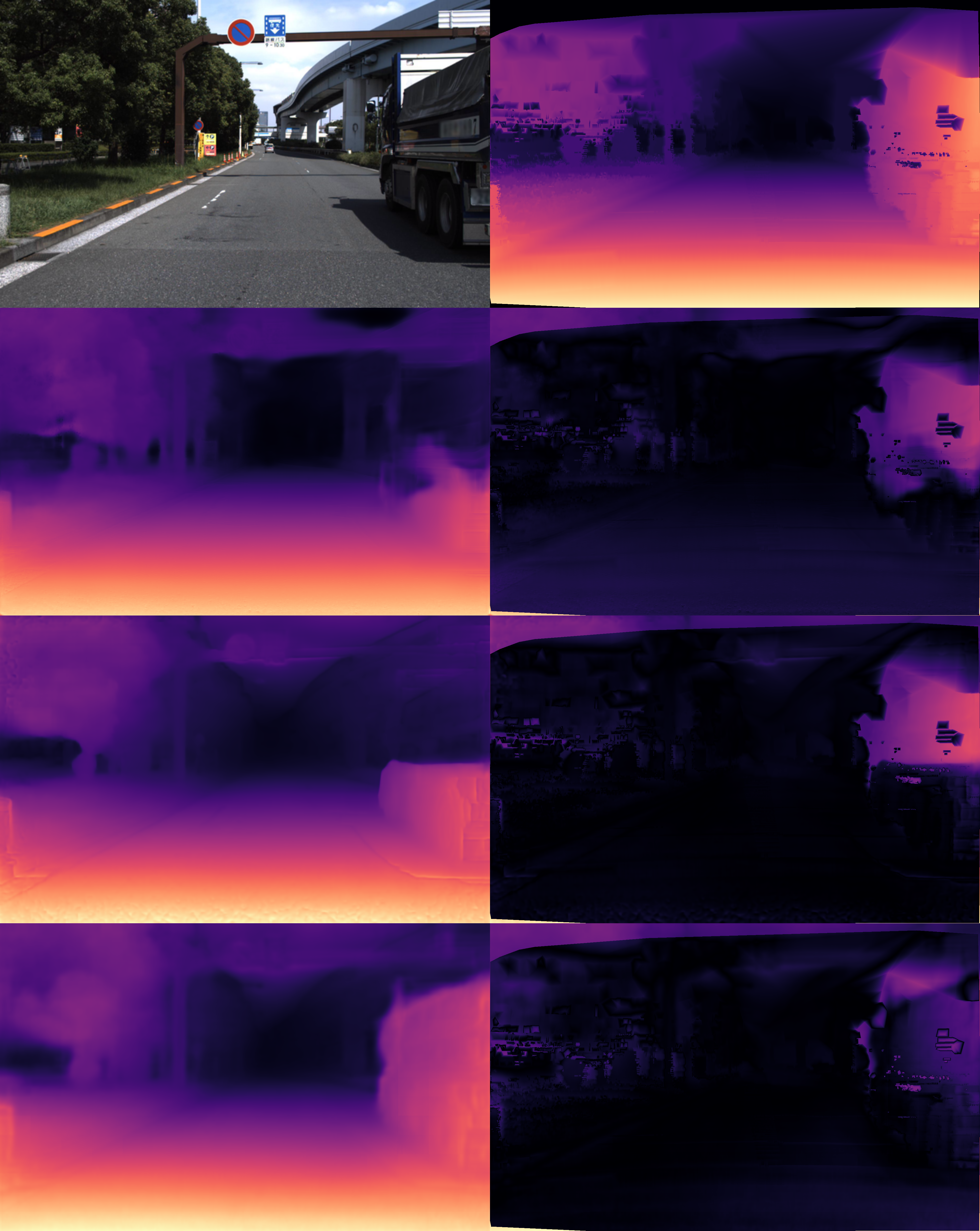}
	\includegraphics[scale=0.03]{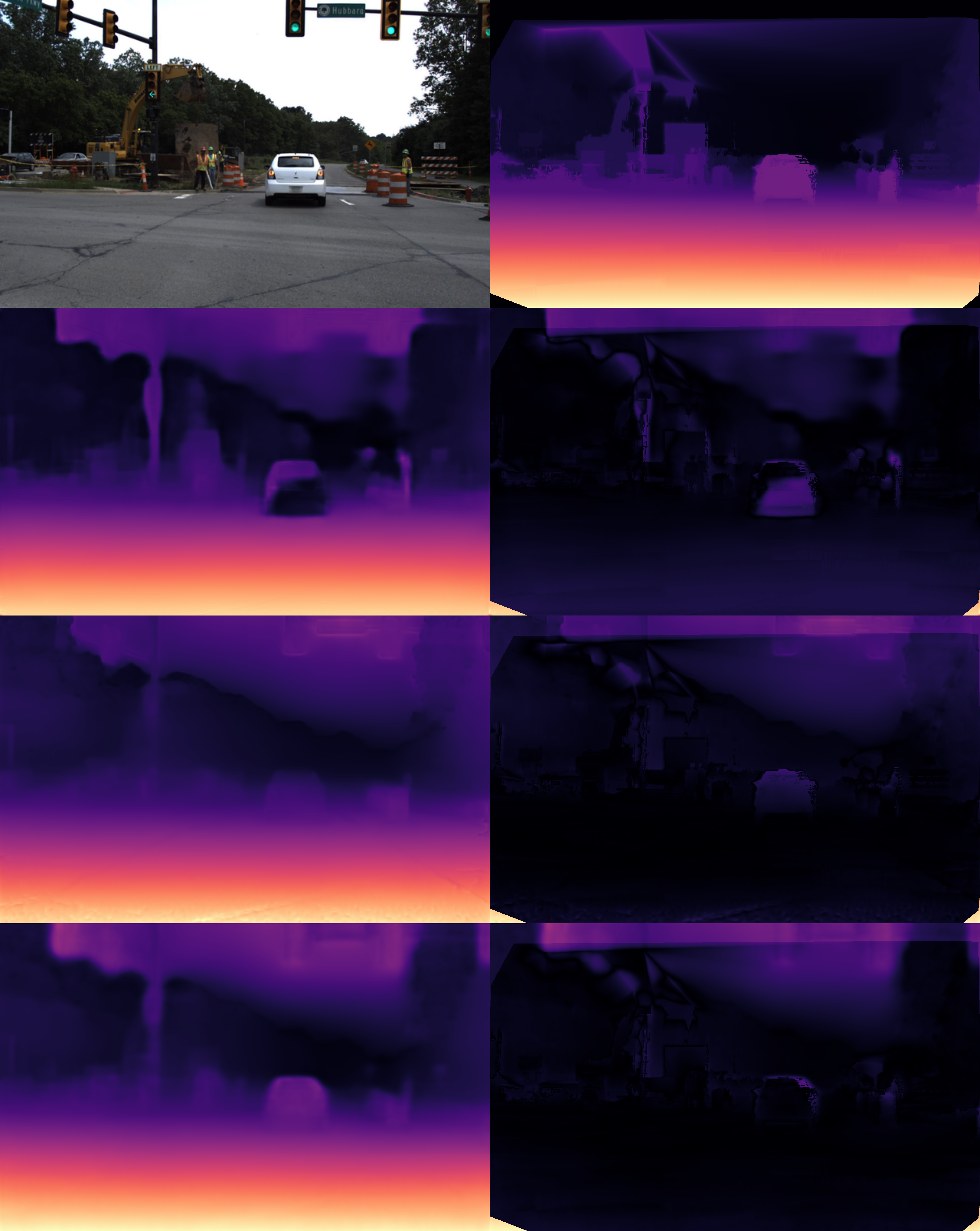}	\includegraphics[scale=0.03]{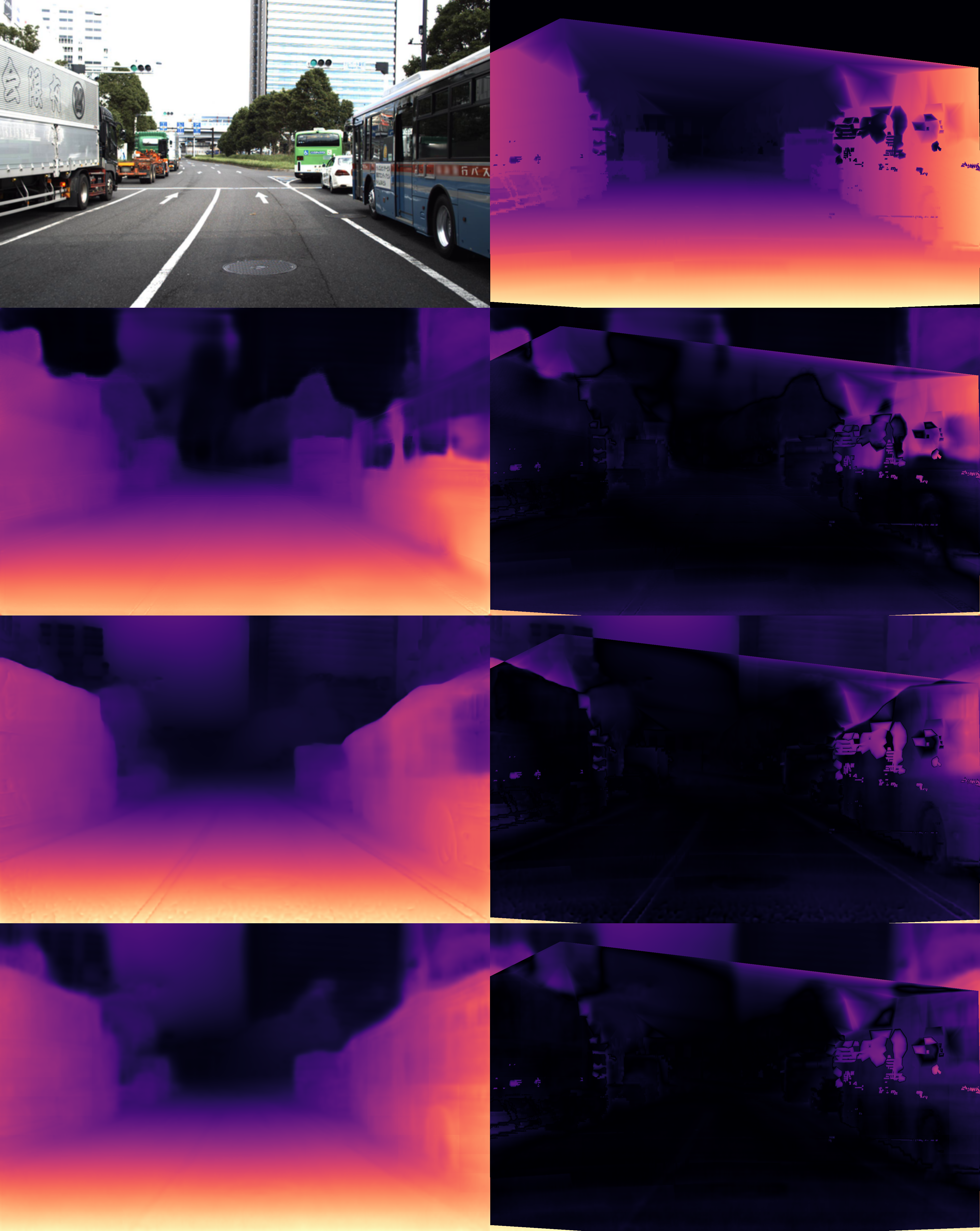}
	\caption{Qualitative comparison among examples of the results obtained on the DDAD \cite{guizilini20203d} dataset. }\label{fig:compare_disp_ddad}%
	\vspace{-10pt}	
\end{figure*}

In Fig. \ref{fig:compare_disp_ddad}, we present the qualitative results obtained by the different methods on the DDAD dataset. A similar phenomenon as that produced on the KITTI dataset can be observed. Examples include the trucks in the left/middle samples, the bus in the right sample, and the moving white car in the middle sample. The previous approaches developed in \cite{guizilini20203d,wang2022cbwloss} inaccurately inferred the depths in these regions, while the proposed HQDec accurately estimated the corresponding depths.

Table \ref{tab:param_time} shows the complexity levels of the tested models. Our method exhibited relatively low computational complexity and can satisfy real-time requirements.

\begin{table}[htbp]\small%[h]%[!htbp] 
	\vspace{-2pt}
	
	\setlength\tabcolsep{2pt}
	\centering

	\begin{tabular}{lcccc} %需要10列
		\toprule %添加表格头部粗线
		\multicolumn{1}{l}{\added{Method}}& \multicolumn{1}{c}{\added{Param(M)}} &\multicolumn{1}{c}{\added{GFLOPs(G)}}& \multicolumn{1}{c}{\added{FPS}} & \multicolumn{1}{c}{\added{GPU-Utili}}\\

		\hline %绘制一条水平横线
		\multicolumn{1}{l}{\added{Zhao et al.}\cite{zhao2022monovit}}&\added{27.87}&\added{13.02}& \added{73} &\added{64\%}\\

		\multicolumn{1}{l}{\added{Zhou et al.}\cite{zhou2021self}}&\added{10.87}&\added{6.84}& \added{69} &\added{67\%}\\
		
		\multicolumn{1}{l}{\added{He et al.}\cite{he2022ra}}&\added{9.98}&\added{4.67}& \added{68} &\added{65\%}\\ 
		\multicolumn{1}{l}{\added{Guizilini et al.}\cite{guizilini20203d}}&\added{128.29}&\added{89.04}& \added{28} &\added{99\%}\\
		
		\multicolumn{1}{l}{\added{Bian et al.}\cite{bian2019unsupervised}}&\added{80.88}&\added{7.81}& \added{121} &\added{90\%}\\
		
		\multicolumn{1}{l}{\added{Godard et al.}\cite{godard2019digging}}&\added{14.84}&\added{3.49}& \added{212} &\added{72\%}\\
		
		\multicolumn{1}{l}{\added{Johnston et al.}\cite{johnston2020self}}&\added{64.45}&\added{56.47}& \added{63} &\added{80\%}\\ 
		
		\multicolumn{1}{l}{\added{Lyu et al.}\cite{lyu2021hr}}&\added{14.61}&\added{7.84}& \added{172} &\added{66\%}\\
		
		\multicolumn{1}{l}{\added{Masoumian et al.}\cite{masoumian2023gcndepth}}&\added{73.85}&\added{13.84}& \added{142} &\added{71\%}\\ 
		
		\multicolumn{1}{l}{\added{Wang et al.}\cite{wang2022cbwloss}}&\added{32.52}&\added{7.22}& \added{145}&\added{70\%} \\ 
		
		\multicolumn{1}{l}{\added{Yan et al.}\cite{yan2021channel}}&\added{58.34}&\added{17.66}& \added{129} &\added{86\%}\\ 
		
		\multicolumn{1}{l}{\added{Ours}}&\added{29.29}&\added{5.90}&\added{36}&\added{51\%}\\  
		
		\bottomrule 
	\end{tabular}
	\caption{\added{Complexity and offline inference time comparison among the different DepthNets with batch size=1. All results were evaluated on an RTX 4090 GPU with the same settings. The resolution was set to 128$\times$416. The time consumption of DepthNet was averaged over 697 test frames in accordance with Eigen’s testing split, and `Param', `GFLOPs', and `FPS' denote the parameter complexity, computational complexity, and frames per second, respectively. `GPU-Utili' represents GPU utilization over most of the time range during the inference process. A smaller value indicates a longer wait time for data during inference. }}\label{tab:param_time}
	\vspace{-5pt}
\end{table}

\begin{table}[htbp]\small%[h]%[!htbp] 
	\vspace{-2pt}
	
	\setlength\tabcolsep{2pt}
	\centering

	\begin{tabular}{lccc} %需要10列
		\toprule %添加表格头部粗线
		\multicolumn{1}{l}{Method}& \multicolumn{1}{c}{RES} &\multicolumn{1}{c}{Seq. 09}& \multicolumn{1}{c}{Seq. 10} \\
		\hline %绘制一条水平横线
		%\multicolumn{1}{l}{ORB-SLAM (short)} &-& 0.064$\pm$0.141 &0.064$\pm$0.130 \\  
		%		 
		%\multicolumn{1}{l}{ORB-SLAM (full)}&-& 0.014$\pm$0.008&0.012$\pm$0.011 \\
	%	\multicolumn{1}{l}{Mean Odometry}&-& 0.032$\pm$0.026& 0.028$\pm$0.023 \\
		%		 
		%\multicolumn{1}{l}{Zhou et al. \cite{zhou2017unsupervised}}&$128 \times 416$& 0.021$\pm$0.017& 0.020$\pm$0.015 \\
		
		\multicolumn{1}{l}{Mahjourian et al. \cite{mahjourian2018unsupervised}}&$128 \times 416$& 0.013$\pm$0.010& 0.012$\pm$0.011 \\
		
		\multicolumn{1}{l}{Jia et al. \cite{jia2021self}}&$128\times 416$ & 0.011$\pm$0.006& \underline{0.009$\pm$0.007} \\
		
	%	\multicolumn{1}{l}{Zou et al. \cite{zou2018df}}&$160\times 576$& 0.017$\pm$0.007& 0.015$\pm$0.009\\

		%192 x640
		\multicolumn{1}{l}{Godard et al. \cite{godard2019digging}}&$192\times 640$ & 0.021$\pm$0.009 & 0.014$\pm$0.010 \\
		
		\multicolumn{1}{l}{Ambrus et al. \cite{ambrus2020two}}&$192 \times 640$& 0.009$\pm$0.004 & \textbf{0.008$\pm$0.007} \\
		
		\multicolumn{1}{l}{Guizilini et al. \cite{guizilini20203d}}&$192\times 640$ & 0.011$\pm$0.006 & \underline{0.009$\pm$0.007} \\
		%\color{blue}
		\multicolumn{1}{l}{\added{He et al. }\cite{he2022ra}}&\added{$192\times 640$}&\added{0.021$\pm$0.009} & \added{0.014$\pm$0.010} \\
		\multicolumn{1}{l}{\added{Zhou et al. }\cite{zhou2021self}}&\added{$192\times 640$}&\added{0.020$\pm$0.009} &\added{0.014$\pm$0.010} \\

		\multicolumn{1}{l}{Bian et al. \cite{bian2019unsupervised}}&$256\times 832$ & 0.016$\pm$0.007 & 0.016$\pm$0.015 \\
		
		\multicolumn{1}{l}{Luo et al. \cite{luo2019every}}&$256 \times 832$& 0.013$\pm$0.007& 0.012$\pm$0.008 \\
		
		\multicolumn{1}{l}{Ranjan et al. \cite{ranjan2019competitive}}&$256 \times 832$& 0.012$\pm$0.007& 0.012$\pm$0.008 \\
		\multicolumn{1}{l}{Wang et al. \cite{wang2022cbwloss}}&$256 \times 832$& 0.012$\pm$0.007& 0.012$\pm$0.008 \\

		\multicolumn{1}{l}{Jia et al. \cite{jia2021self}}&$256 \times 832$& 0.009$\pm$0.007& \underline{0.009$\pm$0.007}\\	
		
		\multicolumn{1}{l}{Wang et al. \cite{wang2022cbwloss}}&$384 \times 1280$& \underline{0.008$\pm$0.005}& \textbf{0.008$\pm$0.006} \\

		\multicolumn{1}{l}{\textbf{Ours}}&$128 \times 416$& \textbf{0.007$\pm$0.004}& \textbf{0.008$\pm$0.007} \\

		\bottomrule 
	\end{tabular}
	\caption{Camera pose prediction performance comparison \added{on short snippets}.}\label{tab:camera_pose_previous_method} 
	\vspace{-10pt}
\end{table}
\begin{figure}[htbp]
\includegraphics[scale=0.25]{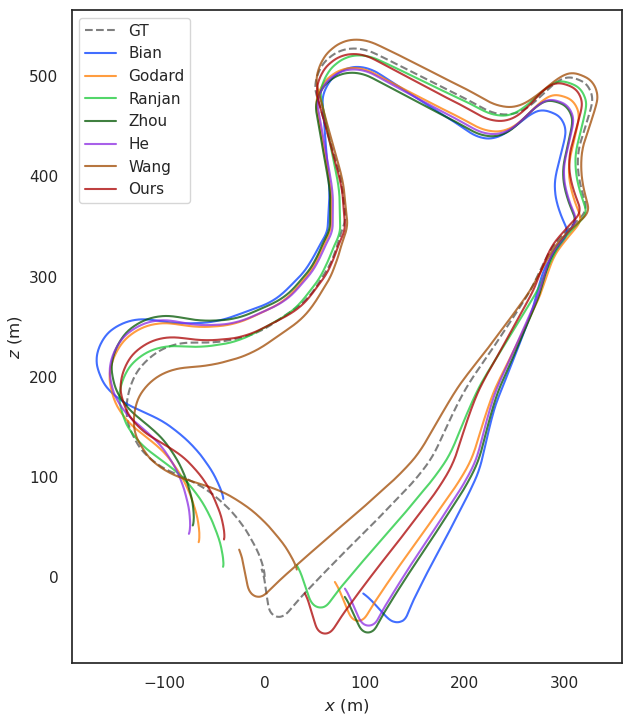}
\includegraphics[scale=0.25]{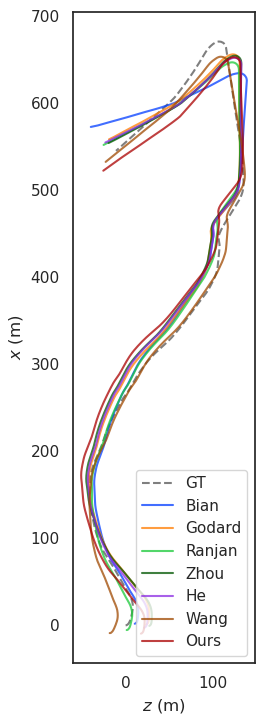}
\caption{\added{Qualitative results obtained on the KITTI odometry dataset without conducting pretraining on the Cityscapes dataset. The resolutions are 128x416/192x640/256x832 for (Ours)/(Godard, Zhou, and He)/(Wang, Ranjan, and Bian). Three frames/five frames snippets were used to train CameraNet for (Ours, Bian, Godard, Zhou, and He)/(Wang, and Ranjan), respectively.}}\label{fig:odo_viz}%
\vspace{-0.7cm}
\end{figure}

\begin{table*}[htbp]
	\centering	
	\resizebox{1\textwidth}{!}{
		\scriptsize		
		\begin{tabular}{lcccccccccccc}
			\toprule%\multirow{2}*{\footnotesize Method}
			\multicolumn{1}{l}{\multirow{2}*{\footnotesize \added{Method}}}&
			\multicolumn{1}{c}{\multirow{2}*{\footnotesize \added{RES}}}&
			\multicolumn{1}{c}{\multirow{2}*{\footnotesize \added{LEN}}} & \multicolumn{3}{c}{\added{Seq. 09}} && \multicolumn{3}{c}{\added{Seq. 10}}
			\\
		
			\multicolumn{3}{c}{}&\added{RMSE (m) }&\added{ Rel. trans. (\%)} &\added{ Rel. rot. (deg/m)} && \added{RMSE (m) }&\added{ Rel. trans. (\%)} & \added{Rel. rot. (deg/m)} \\
			
			\midrule
			\added{Ranjan et al. \cite{ranjan2019competitive}}&\added{$256\times 832$}&\added{5}& \added{\underline{22.82}} & \added{\underline{5.96}} & \added{\underline{0.017}} && \added{13.95} & \added{\textbf{8.06}} & \added{0.032} \\	
			
			\added{Godard et al. \cite{godard2019digging}}&\added{$192\times 640$}&\added{3} &\added{39.83} & \added{8.45} & \added{0.018} && \added{\textbf{12.60}} & \added{\underline{8.46}} & \added{0.033} \\		
			
			\added{He et al. \cite{he2022ra}}&\added{$192\times 640$}& \added{3} & \added{44.82} & \added{9.71} &\added{0.021}&& \added{\underline{13.85}} & \added{9.19} & \added{0.033} \\
			
			\added{Wang et al. \cite{wang2022cbwloss}}&\added{$256\times 832$}& \added{5} & \added{\textbf{16.83}} & \added{6.07} &\added{0.020}&& \added{16.86} & \added{10.77} & \added{\underline{0.027}} \\
			
			\added{Zhou et al. \cite{zhou2021self}}	&\added{$192\times 640$}& \added{3} & \added{46.66} & \added{9.83} &\added{0.021}&& \added{14.19} & \added{9.83} & \added{0.033} \\

			\added{Bian et al. \cite{bian2019unsupervised}}	&\added{$256\times 832$}& \added{3} & \added{56.86} & \added{12.57} &\added{0.033}&& \added{20.71} & \added{10.05} & \added{0.049} \\
			
			\added{\textbf{Ours}}&\added{$128\times 416$}&\added{3}& \added{25.11} & \added{\textbf{5.20}} & \added{\textbf{0.014}} && \added{21.46} & \added{12.31} & \added{\textbf{0.026}} \\
			\bottomrule
		\end{tabular}		
	}\caption{\added{Quantitative comparison among the camera poses for the full trajectories. `LEN' denotes the length of the snippets used to train CameraNet. `RMSE' denotes the root-mean-square error induced on the full trajectories. `Rel. trans.'/`Rel. rot.' denote the average translational error/rotational error, respectively, induced on possible subsequences with lengths of (100, 200,..., 800) meters. The results were derived from GitHub's latest weight values obtained without pretraining on the Cityscapes Dataset.}}
	\label{tab:full_traj}
\end{table*}

\begin{figure*}\centering
	\includegraphics[scale=1.05]{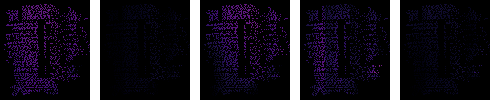}
	 \leftline   {\quad Godard et al. \cite{godard2019digging} \qquad \quad    Guizilini et al. \cite{guizilini20203d}   \qquad \quad Lyu et al. \cite{lyu2021hr} \qquad \qquad Wang et al. \cite{wang2022cbwloss}   \qquad \qquad \qquad   \textbf{Ours}}	
	\caption{\added{Enlarged error map corresponding to the position of the red rectangular box in the left sample of Fig. \ref{fig:compare_disp_kitti}.}}\label{fig:compare_disp_kitti_error_enlarge_60}
	\vspace{-10pt}
\end{figure*}
\begin{figure*}\centering
	\includegraphics[scale=0.82]{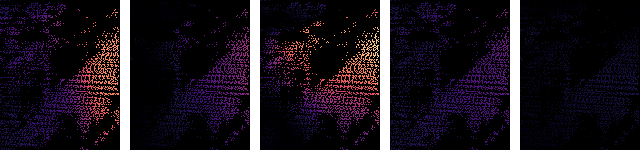}
	\leftline   {\quad Godard et al. \cite{godard2019digging} \qquad \quad    Guizilini et al. \cite{guizilini20203d}   \qquad \quad Lyu et al. \cite{lyu2021hr} \qquad \qquad Wang et al. \cite{wang2022cbwloss}   \qquad \qquad \qquad   \textbf{Ours}}
	\caption{\added{Enlarged error map corresponding to the position of the red rectangular box in the middle sample of Fig. \ref{fig:compare_disp_kitti}.}}\label{fig:compare_disp_kitti_error_enlarge_550}
	\vspace{-10pt}
\end{figure*}
\begin{figure*}\centering
	\includegraphics[scale=0.85]{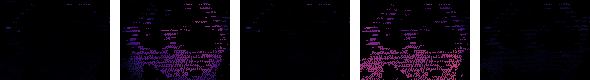}
	\leftline   {\quad Godard et al. \cite{godard2019digging} \qquad \quad    Guizilini et al. \cite{guizilini20203d}   \qquad \quad Lyu et al. \cite{lyu2021hr} \qquad \qquad Wang et al. \cite{wang2022cbwloss}   \qquad \qquad \qquad   \textbf{Ours}}
	\caption{\added{Enlarged error map corresponding to the position of the red rectangular box in the right sample of Fig. \ref{fig:compare_disp_kitti}.}}\label{fig:compare_disp_kitti_error_enlarge_57_car}
	\vspace{-10pt}
\end{figure*}

%   ddad compare

\vspace{-0.3cm}
\subsection{Ablation Studies}

\begin{table*}[htbp]\small%[!htbp] 
	\setlength\tabcolsep{4pt}
	\centering
	%\small

	\begin{tabular}{lccccccccccccccc} %需要10列
		\toprule %添加表格头部粗线
		\multicolumn{1}{l}{\multirow{2}*{\footnotesize Scheme}}&		
		%\multicolumn{1}{c}{\multirow{2}*{\footnotesize Cap (m)}}&
		\multicolumn{1}{c}{\multirow{2}*{\footnotesize Resolutions}} &
		\multicolumn{1}{c}{\multirow{2}*{\footnotesize \added{Total Param}}}&	
		\multicolumn{1}{c}{\multirow{2}*{\footnotesize \added{Enc Param}}}&
		\multicolumn{1}{c}{\multirow{2}*{\footnotesize \added{Dec} Param}}&				   
		%\multicolumn{1}{c}{\multirow{2}*{\footnotesize Seq. 09}}&
		%\multicolumn{1}{c}{\multirow{2}*{\footnotesize Seq. 10}}&
		%\multicolumn{2}{c}{\footnotesize Pose Error$\downarrow$}&	 
		\multicolumn{4}{c}{\footnotesize Depth Error$\downarrow$}& &   
		\multicolumn{3}{c}{\footnotesize Depth Accuracy$\uparrow$}\\
		\multicolumn{5}{c}{}&\footnotesize AbsRel&\footnotesize SqRel&\footnotesize RMSE&\footnotesize RMSE log& &\footnotesize $\delta_1$ &\footnotesize $ \delta_2$&\footnotesize $\delta_3$&	\\ 	 
		%\hline %绘制一条水平横线 
		\midrule
		
		\multicolumn{1}{l}{{\small  BIU}} &128$\times$416&\added{20.22M}&\added{19.85M}&0.37M&0.1164&0.8187&5.0042&0.1915& &0.8519&0.9524&0.9822&\\

		\multicolumn{1}{l}{{\small RCU}} &128$\times$416&\added{24.09M}&\added{19.85M}&4.24M&0.1145&0.8165&4.9564&0.1903& &0.8589&0.9533&0.9818&\\

		\multicolumn{1}{l}{{\small \added{BIU$^{\ast}$}}} &\added{128$\times$416}&\added{28.58M}&\added{19.85M}&\added{8.73M}&\added{0.1129}&\added{0.7948}&\added{5.0990}&\added{0.1932}& &\added{0.8574}&\added{0.9533}&\added{0.9814}&\\

		\multicolumn{1}{l}{{\small NRCU}} &128$\times$416&\added{24.10M}&\added{19.85M}&4.25M&0.1108&0.7531&4.8238&0.1853& &0.8649&0.9562&0.9831&\\

		\multicolumn{1}{l}{{\small AdaNRSU}} &128$\times$416&\added{26.38M}&\added{19.85M}&6.53M&0.1055&0.7299&4.7259&0.1812& &0.8729&0.9584&0.9837&\\

		\multicolumn{1}{l}{{\small DAdaNRSU}} &128$\times$416&\added{28.45M}&\added{19.85M}&8.60M&0.1060&0.7209&4.6367&0.1800& &0.8745&0.9592&0.9841&\\
		
		\bottomrule 
	\end{tabular}
	\caption{Ablation studies involving the upsample module conducted on KITTI. The scale factor was calculated only with the median information. `BIU' indicates that bilinear interpolation-based upsampling was used to recover the resolution of the feature map at each stage in the decoder. \added{`BIU$^{\ast}$' represents stacking two residual modules consisting of five-layer convolutions followed by the ReLU activation function directly after performing bilinear interpolation.} `RCU'/`NRCU'/`AdaNRSU'/`DAdaNRSU' represent the corresponding upsampling modules proposed in Section \ref{upsample_module}. Except for these components, the remainder of the network remained the same in each experiment. }\label{tab:fine_tune_upsample}
	\vspace{-10pt} 
\end{table*}

\begin{table*}[htbp]\small%[!htbp] 
	\setlength\tabcolsep{4pt}
	\centering
	%\small

	\begin{tabular}{lccccccccccccccc} %需要10列
		\toprule %添加表格头部粗线
		\multicolumn{1}{l}{\multirow{2}*{\footnotesize Scheme}}&		
		%\multicolumn{1}{c}{\multirow{2}*{\footnotesize Cap (m)}}&
		\multicolumn{1}{c}{\multirow{2}*{\footnotesize Resolutions}} &
		\multicolumn{1}{c}{\multirow{2}*{\footnotesize \added{Total Param}}}&	
		\multicolumn{1}{c}{\multirow{2}*{\footnotesize \added{Enc Param}}}&
		\multicolumn{1}{c}{\multirow{2}*{\footnotesize \added{Dec} Param}}&		   
		%\multicolumn{1}{c}{\multirow{2}*{\footnotesize Seq. 09}}&
		%\multicolumn{1}{c}{\multirow{2}*{\footnotesize Seq. 10}}&
		%\multicolumn{2}{c}{\footnotesize Pose Error$\downarrow$}&	 
		\multicolumn{4}{c}{\footnotesize Depth Error$\downarrow$}& &   
		\multicolumn{3}{c}{\footnotesize Depth Accuracy$\uparrow$}\\
		\multicolumn{5}{c}{}&\footnotesize AbsRel&\footnotesize SqRel&\footnotesize RMSE&\footnotesize RMSE log& &\footnotesize $\delta_1$ &\footnotesize $ \delta_2$&\footnotesize $\delta_3$&	\\ 	 
		%\hline %绘制一条水平横线 
		\midrule

		\multicolumn{1}{l}{{\small w/o IE}} &128$\times$416&\added{20.22M}&\added{19.85M}&0.37M&0.1164&0.8187&5.0042&0.1915& &0.8519&0.9524&0.9822&\\

		\multicolumn{1}{l}{{\small plain IE }} &128$\times$416&\added{20.50M}&\added{19.85M}&0.65M&0.1153&0.8110&4.9428&0.1905& &0.8526&0.9522&0.9820&\\

		\multicolumn{1}{l}{{\small AdaIE}} &128$\times$416&\added{20.68M}&\added{19.85M}&0.83M&0.1125&0.7885&4.8540&0.1867& &0.8633&0.9556&0.9828&\\

		\bottomrule 
	\end{tabular}
	\caption{Ablation studies involving the information exchange module. The evaluation methods were the same as those in Table \ref{tab:fine_tune_upsample}. `w/o IE' represents that only the connection UNet-based \cite{ronneberger2015u} style was used between the encoder and decoder. The `plain IE' scheme represents the multilevel fine-grained information exchange process without using the learnable parameter tensor. `AdaIE' represents the adaptive information exchange process performed by using the learnable parameter tensor discussed in Section \ref{adaptive_param_module}. Except for these components, the remainder of the network remained the same in each experiment.}\label{tab:adative_infor_exchange_module}
	\vspace{-10pt}  %
\end{table*}

\added{To better understand the contribution of each component proposed in Section \ref{sec:method} to the overall performance of our method, we performed ablation studies on the baseline and different variants of each component in Tables \ref{tab:fine_tune_upsample}, \ref{tab:adative_infor_exchange_module}, \ref{tab:disp_attention_module}, \ref{tab:refine_module}, \ref{tab:downsample_module}, and \ref{tab:scale_alignment}. The differences between the different variants are as follows. Compared with `AdaRM', `RM' in Table \ref{tab:refine_module} directly fuses both the local and global extracted information into the original feature map. Compared with 'DAdaNRSU', 'AdaNRSU' in Table \ref{tab:fine_tune_upsample} models local and long-range dependencies between pixels by utilizing `RM' instead of `AdaRM'. `NRCU' directly fuses the high-frequency information recovered by `RM' into a coarse-grained high-resolution feature map (obtained by bilinear interpolation) by concatenating along the channel dimension. Compared with `NRCU', `RCU' directly fuses the corresponding feature maps without normalizing them. In Table \ref{tab:downsample_module}, compared with `AdaAxialNPCAS', `AdaNPCAS' rearranges the elements in the feature map in a nonaxial way, and recombines the information contained in the rearranged tensor (where the position information of each element in the original feature map is not added) via standard convolution. Compared with `AdaNPCAS', `AdaNCAS' represents each channel by treating each element in the same channel equally.  Compared with `AdaNCAS', `NCAS' performs information fusion directly by concatenating along the channel dimension. Compared with `NCAS', `CAS' only utilizes the channel attention sampling module to obtain a low-resolution feature map.}

The results in Table \ref{tab:fine_tune_upsample} show the impacts of different upsampling schemes on the resulting depth \deleted{and pose} estimation performance. The results demonstrate that the depth \deleted{and pose} estimation effects could be improved by replacing the bilinear interpolation-based upsampling (BIU) process in the decoder with the proposed refined upsampling (RCU) method. \added{However, its parameters also increased. To this end, we increased the model capacity of `BIU' by stacking more convolutions performing after bilinear interpolation in the `BIU' scheme, resulting in the `BIU*' scheme. Although scheme `BIU*' performed better than the `BIU' scheme, it was weaker than the `NRCU' scheme.}\deleted{We attribute the achieved improvement to the fact that RCU can better utilize both the local and long-range dependencies between pixels to recover fine-grained high-resolution feature maps than BIU.} \deleted{It was found that the results (`NRCU' in Table \ref{tab:fine_tune_upsample}) could be further improved by normalizing the high-resolution feature map obtained by BIU and formula \eqref{eq_refine_upsample_pixelshuffle} before being concatenated along the channel dimension.} We believe that this may be because the high-resolution feature maps obtained by \replaced{BIU and formula \eqref{eq_refine_upsample_pixelshuffle}}{the above two schemes} had different data distribution characteristics and semantic levels, making information fusion difficult. The results of the `AdaNRSU' scheme in Table \ref{tab:fine_tune_upsample} show that the depth \deleted{and pose} predictions could be further improved by allowing the model itself to adaptively decide what information contained in the high-resolution feature map (obtained by rearranging the elements in a low-resolution tensor to a high-resolution tensor) needs to be fused into the high-resolution feature map obtained by BIU. Benefitting from the proposed `AdaRM' that could adaptively fuse local and global information, the depth was further improved by replacing the 'RM' module with `AdaRM'. \added{Although `DAdaNRSU' and `BIU' had similar numbers of parameters, `DAdaNRSU' outperformed `BIU*'.}

Table \ref{tab:adative_infor_exchange_module} shows the impacts of different schemes for exchanging information between the multilevel high-resolution feature maps derived from the shallower layers and the feature maps with higher-level semantics on the predicted depth \deleted{and pose}. Compared with using only the UNet-based \cite{ronneberger2015u} style connection (`w/o IE' in Table \ref{tab:adative_infor_exchange_module}) between the encoder and decoder, the proposed `plain IE' scheme could incorporate multilevel spatial information from the shallower layers of different stages into a feature map with higher-level semantics, resulting in better depth \deleted{ and pose} predictions. The results of the `AdaIE' scheme in Table \ref{tab:adative_infor_exchange_module} indicate that the depth \deleted{ and pose} estimation performance can be further improved if the model is allowed to adaptively decide which levels of spatial information from the different stages of the encoder in the shallower layers should be introduced to the feature map with higher-level semantics and how much spatial information should be incorporated into these feature maps. \added{Moreover, compared with `BIU*', which is equivalent to stacking convolutions directly after performing bilinear interpolation in the `w/o IE' scheme with a larger number of  parameters, the `AdaIE' scheme outperformed `BIU*'. }

Table \ref{tab:disp_attention_module} shows the influence of the disparity module on the resulting model performance. The results demonstrate that the performance could be improved if the traditional disparity module (corresponding to the results of `Conv2D Disp' in Table \ref{tab:disp_attention_module}) was assisted with sufficient global statistical properties (corresponding to the results of `AttDisp' in Table \ref{tab:disp_attention_module}). We believe that this may be because sufficient contextual information can provide more semantics for the current pixel, resulting in the model becoming capable of inferring more accurate depths. For example, the learned semantic knowledge can tell the model that the surfaces of the objects in a scene should have more similar depths. \added{Furthermore, compared with `BIU*', which utilizes the traditional disparity module but has a larger number of parameters, `AttDisp' still achieved better depth prediction results.}

\begin{table*}[htbp]\small%[!htbp] 
	\setlength\tabcolsep{4pt}
	\centering
	%\small	
	\begin{tabular}{lccccccccccccccc} %需要10列
		\toprule %添加表格头部粗线
		\multicolumn{1}{l}{\multirow{2}*{\footnotesize Scheme}}&		
		%\multicolumn{1}{c}{\multirow{2}*{\footnotesize Cap (m)}}&
		\multicolumn{1}{c}{\multirow{2}*{\footnotesize Resolutions}} &
		\multicolumn{1}{c}{\multirow{2}*{\footnotesize \added{Total Param}}}&
		\multicolumn{1}{c}{\multirow{2}*{\footnotesize \added{Enc Param}}}&
		\multicolumn{1}{c}{\multirow{2}*{\footnotesize \added{Dec} Param}}&		   
		%\multicolumn{1}{c}{\multirow{2}*{\footnotesize Seq. 09}}&
		%\multicolumn{1}{c}{\multirow{2}*{\footnotesize Seq. 10}}&
		%\multicolumn{2}{c}{\footnotesize Pose Error$\downarrow$}&	 
		\multicolumn{4}{c}{\footnotesize Depth Error$\downarrow$}& &   
		\multicolumn{3}{c}{\footnotesize Depth Accuracy$\uparrow$}\\
		\multicolumn{5}{c}{}&\footnotesize AbsRel&\footnotesize SqRel&\footnotesize RMSE&\footnotesize RMSE log& &\footnotesize $\delta_1$ &\footnotesize $ \delta_2$&\footnotesize $\delta_3$&	\\ 	 
		%\hline %绘制一条水平横线 
		\midrule	%			
		
		\multicolumn{1}{l}{{\small Conv2D disp}} &128$\times$416&\added{20.22M}&\added{19.85M}&0.37M&0.1164&0.8187&5.0042&0.1915& &0.8519&0.9524&0.9822&\\  		
		
		\multicolumn{1}{l}{{\small AttDisp }} &128$\times$416&\added{20.51M}&\added{19.85M}&0.66M&0.1104&0.7643&4.8104&0.1844& &0.8649&0.9565&0.9833&\\		
		\bottomrule 
	\end{tabular}
	\caption{Ablation studies concerning the disparity attention module. The evaluation methods were the same as those in Table \ref{tab:fine_tune_upsample}. `Conv2D disp' represents the disparity regression module (local 2D convolution followed by a sigmoid function) used in previous methods \cite{bian2019unsupervised,ranjan2019competitive,godard2019digging,wang2020unsupervised,wang2022cbwloss,klingner2020self,guizilini2020semantically,guizilini20203d,lyu2021hr,watson2019self,zhou2017unsupervised}. `AttDisp' represents the disparity attention module proposed in Sec. \ref{sec:global_disp_att_module}. Except for these components, the rest of the network remained the same in each experiment. }\label{tab:disp_attention_module}
	\vspace{-10pt}
\end{table*}

\begin{table*}[htbp]\small%[!htbp] 
	\setlength\tabcolsep{4pt}
	\centering
	%\small

	\begin{tabular}{lccccccccccccccc} %需要10列
		\toprule %添加表格头部粗线
		\multicolumn{1}{l}{\multirow{2}*{\footnotesize Scheme}}&		
		%\multicolumn{1}{c}{\multirow{2}*{\footnotesize Cap (m)}}&
		\multicolumn{1}{c}{\multirow{2}*{\footnotesize Resolutions}} &
		\multicolumn{1}{c}{\multirow{2}*{\footnotesize \added{Total Param}}}&	
		\multicolumn{1}{c}{\multirow{2}*{\footnotesize \added{Enc Param}}}&	
		\multicolumn{1}{c}{\multirow{2}*{\footnotesize \added{Dec} Param}}&		   
		%\multicolumn{1}{c}{\multirow{2}*{\footnotesize Seq. 09}}&
		%\multicolumn{1}{c}{\multirow{2}*{\footnotesize Seq. 10}}&
	%	\multicolumn{2}{c}{\footnotesize Pose Error$\downarrow$}&	 
		\multicolumn{4}{c}{\footnotesize Depth Error$\downarrow$}& &   
		\multicolumn{3}{c}{\footnotesize Depth Accuracy$\uparrow$}\\
		\multicolumn{5}{c}{}&\footnotesize AbsRel&\footnotesize SqRel&\footnotesize RMSE&\footnotesize RMSE log& &\footnotesize $\delta_1$ &\footnotesize $ \delta_2$&\footnotesize $\delta_3$&	\\ 	 
		%\hline %绘制一条水平横线 
		\midrule

		\multicolumn{1}{l}{{\small w/o refine}} &128$\times$416&\added{20.50M}&\added{19.85M}&0.65M&0.1153&0.8110&4.9428&0.1905& &0.8526&0.9522&0.9820&\\

		\multicolumn{1}{l}{{\small PRB}} &128$\times$416&\added{21.19M}&\added{19.85M}&1.34M&0.1136&0.8153&5.0026&0.1909& &0.8569&0.9522&0.9812       &\\

		\multicolumn{1}{l}{{\small \added{PRB$^{\ast}$}}} &\added{128$\times$416}&\added{34.97M}&\added{19.85M}&\added{15.12M}&\added{0.1120}&\added{0.7755}&\added{4.8556}&\added{0.1881}& &\added{0.8628}&\added{0.9556}&\added{0.9820}       &\\

		\multicolumn{1}{l}{{\small RM }} &128$\times$416&\added{27.68M}&\added{19.85M}&7.83M&0.1092&0.7714&4.8258&0.1867& &0.8675&0.9553&0.9819&\\

		\multicolumn{1}{l}{{\small AdaRM }} &128$\times$416&\added{29.08M}&\added{19.85M}&9.23M&0.1066&0.7247&4.7484&0.1816& &0.8693&0.9575&0.9841&\\

		\bottomrule 
	\end{tabular}
	\caption{Ablation studies involving the feature map refinement module. The evaluation methods were the same as those in Table \ref{tab:fine_tune_upsample}. `w/o refine' represents that the multilevel fine-grained information was directly fused. `PRB\added{/PRB$^\ast$}' indicates that the fine-grained information at each level was refined by utilizing \added{one/two} residual blocks (two $3\times 3$ convolutions followed by a ReLU function) before performing fusion. `RM'/`AdaRM' represents that the `RM'/`AdaRM' proposed in Sec. \ref{subsec:feat_refine} was used to refine the feature map at each level before performing fusion. Except for these components, the rest of the network remained the same in each experiment.}\label{tab:refine_module}	
	\vspace{-10pt}
\end{table*}
\begin{table*}[htbp]\small%[!htbp] 
	\setlength\tabcolsep{4pt}
	\centering
	%\small

	\begin{tabular}{lccccccccccccccc}
		\toprule %添加表格头部粗线
		\multicolumn{1}{l}{\multirow{2}*{\footnotesize Scheme}}&		
		%	\multicolumn{1}{c}{\multirow{2}*{\footnotesize Cap (m)}}&
		\multicolumn{1}{c}{\multirow{2}*{\footnotesize Resolutions}} &
		\multicolumn{1}{c}{\multirow{2}*{\footnotesize \added{Total Param}}}&	
		\multicolumn{1}{c}{\multirow{2}*{\footnotesize \added{Enc Param}}}&
		\multicolumn{1}{c}{\multirow{2}*{\footnotesize \added{Dec} Param}}&	   
		%\multicolumn{1}{c}{\multirow{2}*{\footnotesize Seq. 09}}&
		%\multicolumn{1}{c}{\multirow{2}*{\footnotesize Seq. 10}}&
		%	\multicolumn{2}{c}{\footnotesize Pose Error$\downarrow$}&	
		
		\multicolumn{4}{c}{\footnotesize Depth Error$\downarrow$}&   
		\multicolumn{3}{c}{\footnotesize Depth Accuracy$\uparrow$}\\
		\multicolumn{5}{c}{}&\footnotesize AbsRel&\footnotesize SqRel&\footnotesize RMSE&\footnotesize RMSElog& \footnotesize $\delta_1$ &\footnotesize $ \delta_2$&\footnotesize $\delta_3$&	\\ 	 
		%\hline %绘制一条水平横线 
		\midrule
	
		\multicolumn{1}{l}{{\small maxpooling}} &128$\times$416&\added{20.34M}&\added{19.85M}& 0.49M&0.1158&0.8051&4.9501&0.1909 &0.8513&0.9520&0.9818   &\\

		\multicolumn{1}{l}{{\small stride}} &128$\times$416&\added{20.97M}&\added{19.85M}&1.12M&0.1157&0.8170&5.0172&0.1909& 0.8528&0.9518&0.9820&\\

		\multicolumn{1}{l}{{\small maxpooling+stride}} &128$\times$416&\added{20.50M}&\added{19.85M}&0.65M&0.1153&0.8110&4.9428&0.1905& 0.8526&0.9522&0.9820&\\   %  

		\multicolumn{1}{l}{{\small 3D packing}} &128$\times$416&\added{28.32M}&\added{19.85M}&8.47M&0.1129&0.8024&4.9460&0.1893& 0.8587&0.9531&0.9817&\\ %

		\multicolumn{1}{l}{{\small CAS}} &128$\times$416&\added{79.10M}&\added{19.85M}&59.25M&0.1080&0.7491&4.7819&0.1833& 0.8701&0.9575&0.9831   &\\

		\multicolumn{1}{l}{{\small NCAS}} &128$\times$416&\added{35.04M}&\added{19.85M}&15.19M&0.1062&0.7590&4.7246&0.1832& 0.8723&0.9574&0.9829&\\

		\multicolumn{1}{l}{{\small AdaNCAS}} &128$\times$416&\added{79.27M}&\added{19.85M}&59.42M&0.1051&0.7481&4.7107&0.1819& 0.8758&0.9580&0.9832&\\

		\multicolumn{1}{l}{{\small AdaNPCAS}} &128$\times$416&\added{79.28M}&\added{19.85M}&59.43M&0.1048&0.7332&4.6057&0.1792& 0.8775&0.9600&0.9839&\\

		\multicolumn{1}{l}{{\small AdaAxialNPCAS}} &128$\times$416&\added{21.46M}&\added{19.85M}&1.61M&0.1043&0.7269&4.6585&0.1802& 0.8760&0.9591&0.9836&\\

		\bottomrule %添加表格底部粗线
	\end{tabular}
	\caption{Ablation studies concerning downsampling. The evaluation methods were the same as those in Table \ref{tab:fine_tune_upsample}. `3D packing' represents the downsampling process in  \cite{guizilini20203d}. `CAS'/` NCAS'/`AdaNCAS'/`AdaNPCAS'/`AdaAxialNPCAS' represents the  downsampling scheme proposed in Sec. \ref{channel_attention_sample}. Except for these components, the rest of the network remained the same in each experiment.}\label{tab:downsample_module}
	\vspace{-10pt}
\end{table*}

\begin{figure}\centering
	\rotatebox{90}{ \quad {\footnotesize\textbf{ Ours}}\;\qquad
		{\footnotesize Wang \cite{wang2022cbwloss}}\quad  \quad
		{\footnotesize Lyu \cite{lyu2021hr}}\quad  \quad
		{\footnotesize Guizilini \cite{guizilini20203d}}\quad
		{\footnotesize Godard \cite{godard2019digging}} 		
		%{\footnotesize Input }
	}
	\includegraphics[scale=0.65]{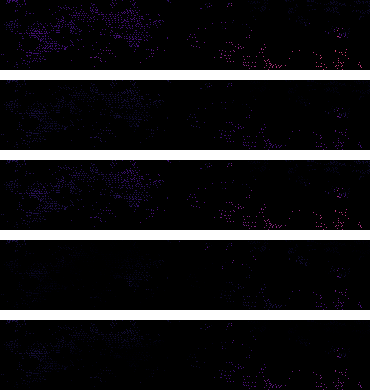}
	\caption{\added{Enlarged error map corresponding to the position of the red rectangular box in the right sample of Fig. \ref{fig:compare_disp_kitti}.}}\label{fig:compare_disp_kitti_error_enlarge_57_lawn}
	\vspace{-10pt}
\end{figure}
\begin{table*}[htbp] \small%[!htbp] 
	\setlength\tabcolsep{4pt}
	\centering
	%\small	
	\begin{tabular}{lccccccccccccccc} %需要10列
		\toprule %添加表格头部粗线
		\multicolumn{1}{l}{\multirow{2}*{\footnotesize Scheme}}&		
		%	\multicolumn{1}{c}{\multirow{2}*{\footnotesize Cap (m)}}&
		\multicolumn{1}{c}{\multirow{2}*{\footnotesize Resolutions}} &
		\multicolumn{1}{c}{\multirow{2}*{\footnotesize \added{Total Param}}}&
		\multicolumn{1}{c}{\multirow{2}*{\footnotesize \added{Enc Param}}}&
		\multicolumn{1}{c}{\multirow{2}*{\footnotesize \added{Dec} Param}}&		   
		%\multicolumn{1}{c}{\multirow{2}*{\footnotesize Seq. 09}}&
		%\multicolumn{1}{c}{\multirow{2}*{\footnotesize Seq. 10}}&
		%\multicolumn{2}{c}{\footnotesize Pose Error$\downarrow$}&	
		
		\multicolumn{4}{c}{\footnotesize Depth Error$\downarrow$}&   
		\multicolumn{3}{c}{\footnotesize Depth Accuracy$\uparrow$}\\
		\multicolumn{5}{c}{}&\footnotesize AbsRel&\footnotesize SqRel&\footnotesize RMSE&\footnotesize RMSElog& \footnotesize $\delta_1$ &\footnotesize $ \delta_2$&\footnotesize $\delta_3$&	\\ 	 
		%\hline %绘制一条水平横线 
		\midrule
	
		\multicolumn{1}{l}{{\small median ($\zeta=1$)}} &128$\times$416&\added{29.29M}&\added{19.85M}& 9.44M&0.1026&0.7063&4.5689&0.1763 &0.8822&0.9615&0.9845   &\\

		\multicolumn{1}{l}{{\small mean ($\zeta=0$)}} &128$\times$416&\added{29.29M}&\added{19.85M}& 9.44M&0.1158&0.7079&4.3250&0.1743 &0.8813&0.9649&0.9862   &\\ 
		
		\multicolumn{1}{l}{{\small fuse ($\zeta=0.5$)}} &128$\times$416&\added{29.29M}&\added{19.85M}& 9.44M&0.1051&0.6870&4.3986&0.1720 &0.8874&0.9644&0.9857   &\\ 
		
		\multicolumn{1}{l}{{\small AdaSearch}} &128$\times$416&\added{29.29M}&\added{19.85M}& 9.44M&0.0994&0.6926&4.4937&0.1732 &0.8866&0.9627&0.9849   &\\

		\bottomrule %添加表格底部粗线
	\end{tabular}
	\caption{Ablation studies involving different scale alignment strategies. `AdaSearch' represents that the $\zeta$ parameter of formula \eqref{eq_ada_fuse} was adaptively searched. The same decoder consisting of `DAdaNRSU', `AdaIE', `AttDisp', `AdaRM', and `AdaAxialNPCAS' was used.  }\label{tab:scale_alignment}	
	\vspace{-10pt}
\end{table*}
\begin{table}[htbp]\small%[h]%[!htbp] 
	\vspace{-2pt}
	
	\setlength\tabcolsep{2pt}
	\centering

	\begin{tabular}{lcccccc} %需要10列
		\toprule %添加表格头部粗线
		\multicolumn{1}{l}{\added{BackBone}}& \multicolumn{1}{l}{\added{\footnotesize Resolutions}}&\multicolumn{1}{c}{\added{AbsRel}} &\multicolumn{1}{c}{\added{SqRel}}& \multicolumn{1}{c}{\added{RMSE}}& \multicolumn{1}{c}{\added{RMSElog}}&\multicolumn{1}{c}{\added{$\delta_1$}} \\
		\hline %绘制一条水平横线		
		\multicolumn{1}{l}{\added{RN50+PN7}}&\added{$128\times 416$}&\added{0.1052}&\added{0.7797}& \added{4.7137}&\added{0.1848} &\added{0.8813}\\
		
		\multicolumn{1}{l}{\added{Effv2s+FBv3}}&\added{$128\times 416$}&\added{0.1026}&\added{0.7063}& \added{4.5689}&\added{0.1763} &\added{0.8822}\\

		\bottomrule %添加表格底部粗线  
	\end{tabular}
	\caption{\added{Performance comparison on the different backbone with the proposed decoder HQDec. The scale factor was calculated only with the median information. `RN50/Effv2s' denotes the ResNet50\cite{he2016deep}/EfficientNet2-s\cite{tan2021efficientnetv2} is used as the encoder for DepthNet. `PN7/FBv3' refers to  a simple pose estimation network\cite{wang2022cbwloss}/FBNetV3-B\cite{dai2021fbnetv3} is used as the CameraNet. }}\label{tab:different_backbone}
	\vspace{-22pt}
\end{table}

Table \ref{tab:refine_module} shows the impacts of different refinement schemes on the depth \deleted{and pose} estimation results. Compared with `w/o IE' in Table \ref{tab:adative_infor_exchange_module}, `w/o refine' in Table \ref{tab:refine_module} (namely, `plain IE' in Table \ref{tab:adative_infor_exchange_module}) could yield improved performance, but this improvement was limited because the features derived from different stages had different semantics. To this end, we designed \deleted{three} different modules for refining the feature map to mitigate the above effect before performing direct fusion. The results shown in Table \ref{tab:refine_module} indicate that this method could help with the information fusion process and improve the prediction results if the feature maps derived from different stages were refined before being fused. \added{`PRB*' shows that increasing the capacity of the refinement module could further improve the depth predictions, which occurred because `PRB*' has a larger capacity and receptive field, which could transform feature maps derived from different stages into feature maps with more similar semantics than those of `PRB'.} Benefitting from its ability to capture local information and model long-range dependencies in parallel, the proposed `RM' scheme outperformed `PRB' and \added{even `PRB*', which has a larger capacity}.  Moreover, because the feature maps generated by the local filter and transformer had different semantics, the proposed `AdaRM', in which the feature maps could be adaptively fused by utilizing the learned parameter tensor, was more beneficial for information fusion than `RM'.

Table \ref{tab:downsample_module} shows the impacts of different downsampling modules on the resulting model performance. The results indicate that the traditional downsampling schemes (e.g., max pooling and strided convolution) achieved the worst performance among the downsampling schemes in Table \ref{tab:downsample_module}, which may be due to the loss of information induced by downsampling. Combining max pooling-based downsampling with strided convolution could mitigate these effects. Compared with max pooling and/or strided convolution, the 3D packing \cite{guizilini20203d} scheme could propagate and preserve more details, resulting in the recovery of accurate depths. The proposed `CAS' method directly modeled the dependencies between the expanded structure information by learning cross-channel interactions via the utilization of lightweight 1D convolutions, resulting in better depth predictions than those of 3D packing. It was found that the depth \deleted{and pose} prediction performance could be further improved by fusing the saliency information obtained via max pooling into the above fine-grained representations, namely, the `NCAS' and `AdaNCAS' methods shown in Table \ref{tab:downsample_module}. Moreover, compared with the proposed `NCAS' method, where the saliency information and the fine-grained representations were fused by implementing concatenation in the channel dimension after performing normalization, the proposed `AdaNCAS' is more conducive to information fusion because this scheme allows the model to decide what information needs to be propagated to the decoder. Compared with the `AdaNCAS' scheme that directly calculated each piece of  channel information by 
treating each element of the feature map equally, the proposed `AdaNPCAS', which could focus on the most important elements and give these elements greater weights through an elementwise learnable parameter tensor with an initial value of one, could achieve better performance. The proposed `AdaAxialNPCAS' approach could achieve performance that was similar to or even better than that of the proposed `AdaNPCAS' method in terms of some metrics, with 37 times fewer parameters. \added{The proposed `AdaAxialNPCAS' (1.61 M  parameters) outperformed the 3D packing \cite{guizilini20203d}(8.47 M  parameters) and `PRB*' (15.12 M parameters) methods that utilize residual blocks to process the downsampled (`max pooling+stride') feature maps from different stages, which is attributed to the fact that the `AdaAxialNPCAS' architecture could propagate and preserve more details. }

In Table \ref{tab:scale_alignment}, we investigate the effects of different scale alignment strategies on the obtained absolute depths. The results show that the absolute depths obtained with the median information concerning some indices (e.g., `AbsRel', `SqRel', and `$\delta_1$') were better than those obtained with mean information. For the other indices (e.g., `RMSE', RMSElog', `$\delta_2$, and `$\delta_3$'), the opposite phenomenon was observed. As a simple compromise in which the mean and median information were equally considered, namely, `fuse', the obtained absolute depths for the above indices except `AbsRel' were better than those of the techniques that only considered median information. Compared with `fuse', `AdaSearch' could adaptively select the appropriate scale factor from the scale search space, resulting in more accurate absolute depths.

\section{Conclusions}\label{sec:conclusions}

In this paper, we design an HQDec for DepthNet. Our experimental results indicate that the proposed method outperforms previously developed state-of-the-art single-frame depth estimation methods, including those that use semantic labels as guidance signals. Although we only use single-frame information during the inference process, our method equals or even exceeds methods that use multiframe image information during inference.
However, we observe that the `black hole' problem, which arises from the presence of moving objects in a scene that violates the static scene assumption, still exists. Additionally, during training, our method still requires known camera intrinsics, which forbids the use of random internet videos with unknown camera types. \added{Slight grid artifacts, which are derived from the adopted global dependence modeling scheme, may occur in some areas of the predicted depth. The parameters of the model vary slightly depending on the resolution of the input image.} We plan to address these problems in future work.

\vspace{-0.3cm}
{
	\bibliographystyle{IEEEtran}
	\bibliography{reference}
	
} 
\vspace{-1.8cm}
\begin{IEEEbiography}
	[{\includegraphics[width=0.72in,height=0.9in,clip]{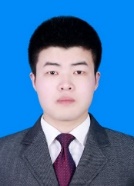}}]	
	{Fei Wang} is currently pursuing the Ph.D. degree in University of Chinese Academy of Sciences, Shenzhen Institute of Advanced Technology. His current research interests include computer vision, structure from motion, robotics and deep learning.
\end{IEEEbiography}
\vspace{-2cm}
\begin{IEEEbiography}
	[{\includegraphics[width=0.72in,height=0.9in,clip]{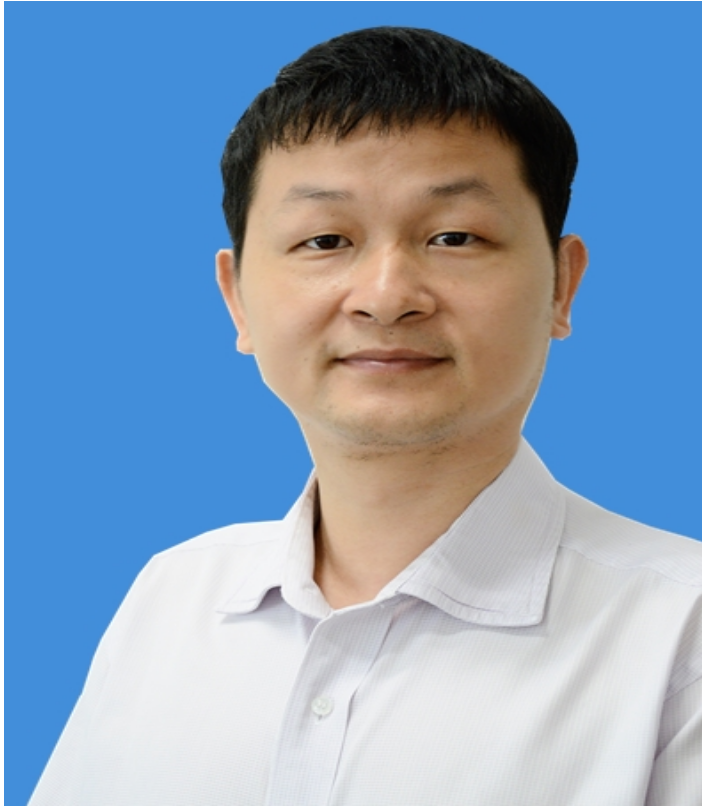}}]
	{Jun Cheng} received the B.Eng. and M.Eng. degrees from the University of Science and Technology of China, Hefei, China, in 1999 and 2002, respectively, and the Ph.D. degree from The Chinese University of Hong Kong, Hong Kong, in 2006. He is currently with the Shenzhen Institute of Advanced Technology, Chinese Academy of Sciences, Shenzhen, China, as a Professor and the Director of the Laboratory for Human Machine Control. His current research interests include computer vision, robotics, machine intelligence, and control.
\end{IEEEbiography}

\end{document}